\documentclass[pdflatex,sn-mathphys-num]{sn-jnl}

\usepackage{graphicx}%
\usepackage{multirow}%
\usepackage{amsmath,amssymb,amsfonts}%
\usepackage{amsthm}%
\usepackage{mathrsfs}%
\usepackage[title]{appendix}%
\usepackage{xcolor}%
\usepackage{textcomp}%
\usepackage{manyfoot}%
\usepackage{booktabs}%
\usepackage{algorithm}%
\usepackage{algorithmicx}%
\usepackage{algpseudocode}%
\usepackage{listings}%
\usepackage{caption}
\usepackage{subcaption}
\usepackage{makecell}

\theoremstyle{thmstyleone}%
\newtheorem{theorem}{Theorem}
\newtheorem{proposition}[theorem]{Proposition}%

\theoremstyle{thmstyletwo}%

\theoremstyle{thmstylethree}%

\begin{document}

\title[Sissi: Zero-shot \underline{S}tyle-guided \underline{I}mage \underline{S}ynthesis via \underline{S}emantic–style \underline{I}ntegration]{Sissi: Zero-shot \underline{S}tyle-guided \underline{I}mage \underline{S}ynthesis via \underline{S}emantic–style \underline{I}ntegration}


\author[1]{\fnm{Yingying} \sur{Deng}}\email{ yingying.deng@ustb.edu.cn}

\author*[2]{\fnm{Xiangyu} \sur{He}}\email{hexiangyu17@mails.ucas.edu.cn}

\author[3]{\fnm{Fan} \sur{Tang}}\email{tfan.108@gmail.com}
\author[2]{\fnm{Weiming} \sur{Dong}}\email{weiming.dong@ia.ac.cn}
\author[1]{\fnm{Xucheng} \sur{Yin}}\email{xuchengyin@ustb.edu.cn}
\affil[1]{\orgdiv{Department of Computer Science and Technology}, \orgname{University of Science and Technology Beijing}, \orgaddress{\city{Beijing}, \postcode{100083}, \state{China}}}

\affil*[2]{\orgdiv{Institute of Automation}, \orgname{Chinese Academy of Sciences}, \orgaddress{ \city{Beijing}, \postcode{100190}, \state{China}}}

\affil[3]{\orgdiv{School of Computer Science and Technology}, \orgname{ University of Chinese Academy of Sciences}, \orgaddress{ \city{Beijing}, \postcode{100040}, \state{China}}}


\abstract{Text-guided image generation has advanced rapidly with large-scale diffusion models, yet achieving precise stylization with visual exemplars remains difficult. Existing approaches often depend on task-specific retraining or expensive inversion procedures, which can compromise content integrity, reduce style fidelity, and lead to an unsatisfactory trade-off between semantic prompt adherence and style alignment.
In this work, we introduce a training-free framework that reformulates style-guided synthesis as an in-context learning task. Guided by textual semantic prompts, our method concatenates a reference style image with a masked target image, leveraging a pretrained ReFlow-based inpainting model to seamlessly integrate semantic content with the desired style through multimodal attention fusion. 
We further analyze the imbalance and noise sensitivity inherent in multimodal attention fusion and propose a Dynamic Semantic-Style Integration (DSSI) mechanism that reweights attention between textual semantic and style visual tokens, effectively resolving guidance conflicts and enhancing output coherence.
Experiments show that our approach achieves high-fidelity stylization with superior semantic-style balance and visual quality, offering a simple yet powerful alternative to complex, artifact-prone prior methods.}




\maketitle

\section{Introduction}
\label{sec:Introduction}

\begin{figure*}
\centering
\includegraphics[width=1.0\linewidth]{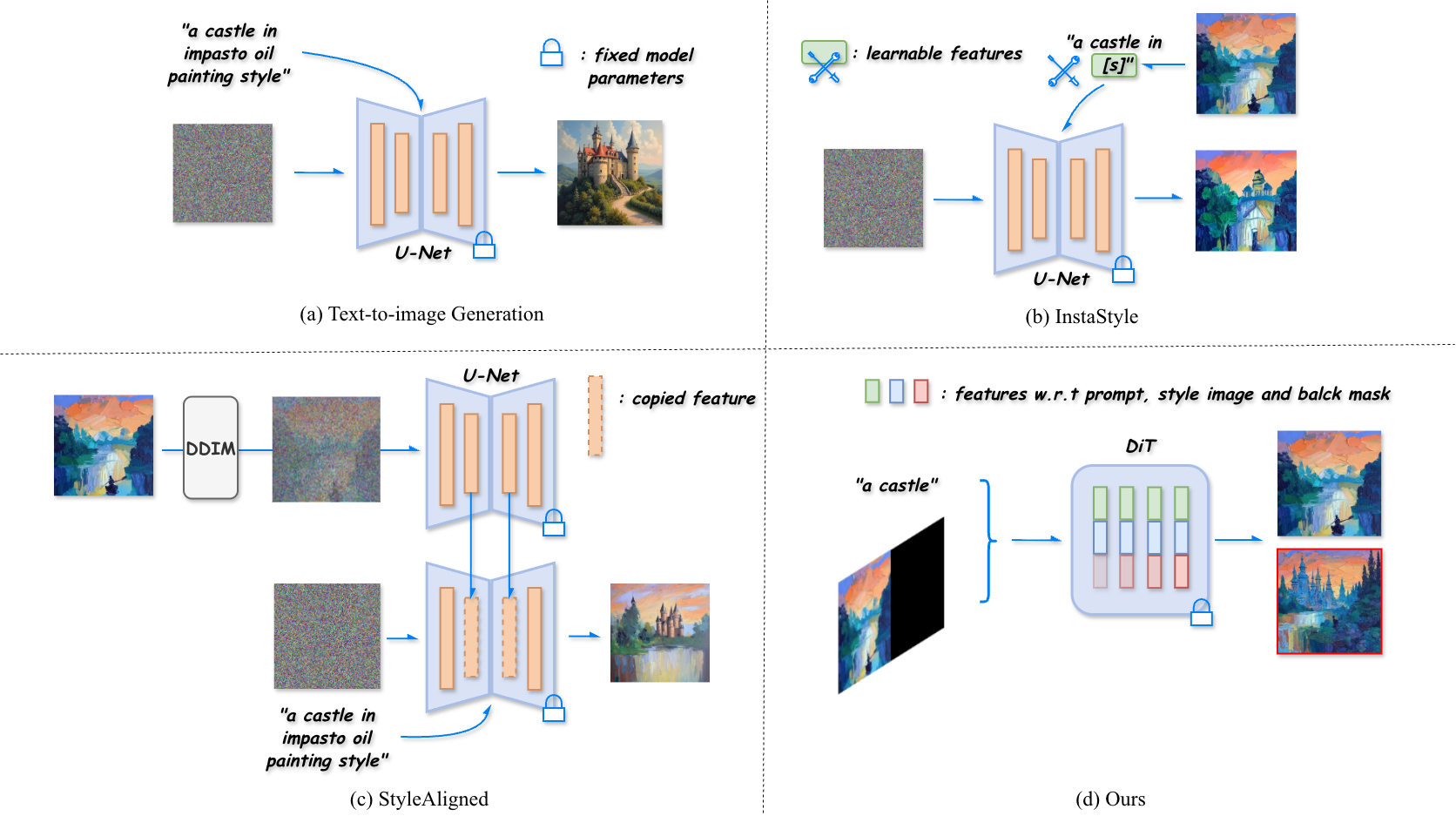}
\caption{Different pipelines for style-guided image synthesis: (a) style-related text-guided image synthesis; (b) training-based style-guided image synthesis (InstaStyle~\cite{INSTASTYLE}); (c) training-free style-guided image synthesis (StyleAligned~\cite{Style_Aligned}); and (d) our method, which simply leverages in-context learning in a pre-trained ReFlow inpainting model enhanced by dynamic semantic-style integration for robust fusion.}
\label{fig:intro}
\end{figure*}

\begin{figure*}
\centering
\includegraphics[width=1.0\linewidth]{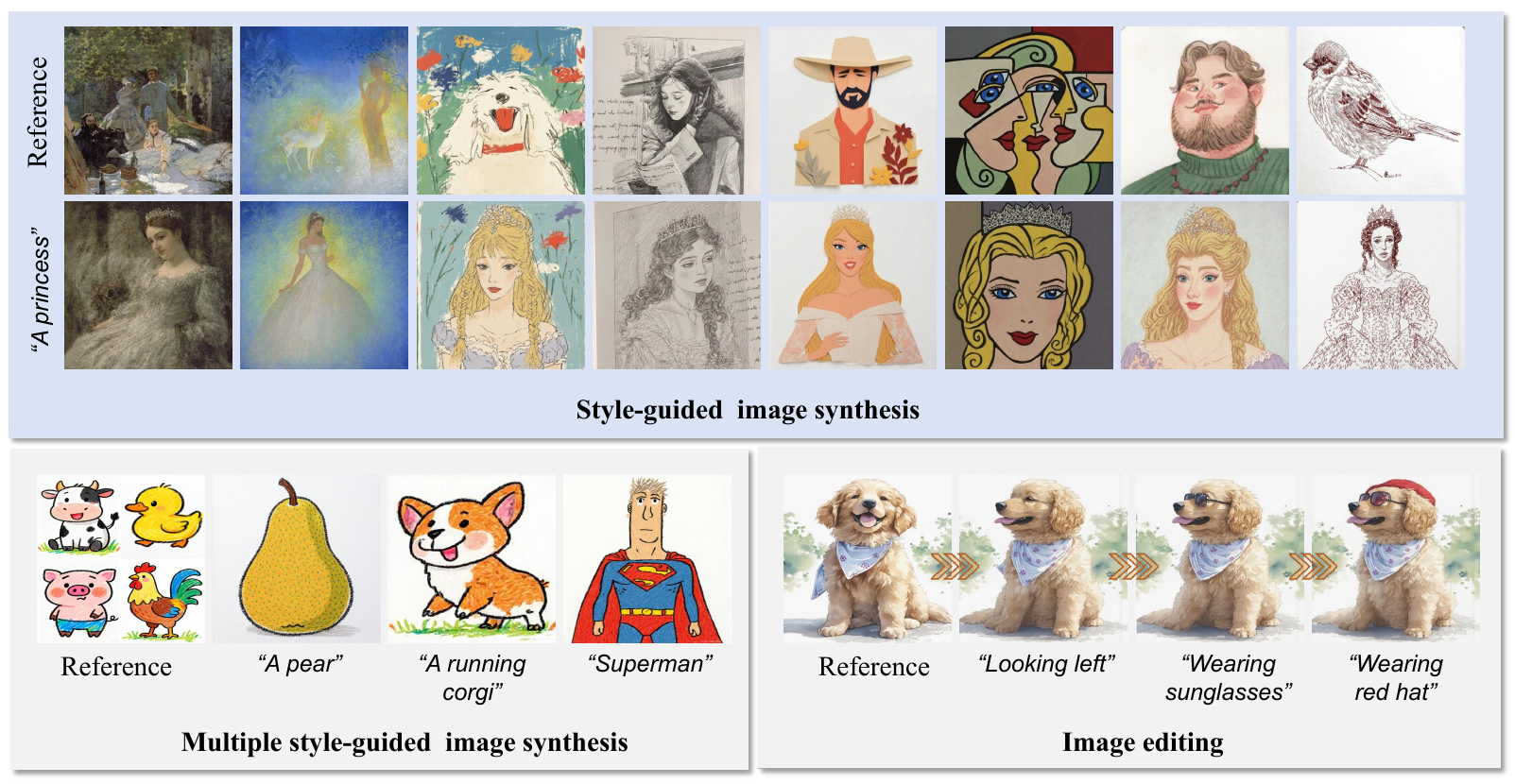}
\caption{Examples of style-guided image synthesis using our method \textbf{Sissi}. Given a style reference, our approach generates stylized images conditioned on a content prompt, achieving exceptional quality and fidelity. Furthermore, it seamlessly extends to multiple style-guided synthesis and image editing tasks, demonstrating versatility and superior performance.
}
\label{fig:p1}
\end{figure*}
Text-guided image generation has gained increasing attention due to the rise of large-scale diffusion models, which allow users to specify desired styles through textual prompts. However, relying solely on text for style guidance presents several limitations. As shown in Fig.~\ref{fig:intro}(a), the abstractness and ambiguity of textual style descriptions make it difficult to capture fine-grained visual details such as brush strokes, textures, subtle color variations, material gloss, or specific artistic effects (e.g., impasto in oil paintings). Moreover, a short prompt often cannot fully convey these details, potentially resulting in style transfer outputs that deviate from user expectations. 

To overcome these limitations, recent research has incorporated style reference images to provide explicit visual examples of target styles. 
A set of methods~\cite{zhang:2023:inversion, INSTASTYLE} directly learn the corresponding textual embedding from images, using the expressive power of the textual space to convey the stylistic characteristics of artistic images.
Adapter-based fine-tuning for text-to-image diffusion models~\cite{T2I_Adapter, ArtAdapter} integrate style control through a learned adapter module, and encode the style image to style textual embeddings which is also concatenated to text embedding to guide the stylized results.
However, the image-guided approaches mentioned above typically require task-specific training and exhibit limited generalization to unseen styles. Moreover, their tendency to entangle structural details from the reference style image often leads to undesired content distortions and overfitting (see Fig.~\ref{fig:intro}(b)).
Alternatively, training-free inversion-based methods~\cite{StyleID, StyleShot, Style_Aligned} leverage DDIM inversion~\cite{ddim_paper} and latent feature fusion to inject visual style from a reference image into the text-conditioned latent space. While avoiding retraining, these approaches are computationally expensive during inference and sensitive to inaccurate attention estimation, often producing artifacts or unstable style alignment (see Fig.~\ref{fig:intro}(c)). Collectively, existing approaches struggle to simultaneously achieve high visual quality, strong style–content consistency and robustness.

Recent progress in Rectified Flow (ReFlow) models provides a potential direction.
Built upon Diffusion Transformer (DiT) backbones, ReFlow models substitute conventional U-Net convolutions with transformer-based sequence modeling, effectively capturing long-range dependencies across spatial and stylistic dimensions. This architectural design endows the model with exceptional in-context learning (ICL) capabilities~\cite{lhhuang2024iclora}, facilitating rapid adaptation to multimodal conditioning signals. However, existing ICL-based image generation tasks~\cite{In-context1, In-context2} typically rely on explicit exemplar pairs for guidance, making it challenging to decouple appropriate style patterns from a single style exemplar and integrate them with a target semantic prompt in ambiguous scenarios.

In this paper, we propose a novel training-free style-guided text-to-image synthesis method that reformulates the problem as an in-context style learning task. Conditioned on textual semantic prompts, our method concatenates a reference style image with a masked target image to create a structured input, which is then fed into a pre-trained ReFlow-based inpainting model. This design enables the text prompt to provide high-level semantic and content guidance, while the style image supplies concrete visual exemplars of textures, colors, and artistic characteristics, facilitating spatially intuitive integration between style and content regions.
However, this multimodal fusion introduces core technical difficulties, including imbalance and noise sensitivity in attention mechanisms. Concatenated attention often results in dominance by stronger activations, such as texture-rich styles overwhelming semantics, and suppression of the weaker branch. Noise introduced by the zero-mask in the inpainting pipeline further exacerbates this issue, leading to amplified perturbation across sampling steps.
To address these problems and ensure coherent integration of semantic and stylistic signals, we introduce a semantic-style integration mechanism that dynamically reweights attention between textual and visual tokens, adaptively resolving conflicts between the guidance sources. The entire framework operates in a training-free manner, thereby enabling effective generalization to unseen styles.

In summary, our key contributions are as follows:
\begin{itemize}
    \item We propose a novel training-free framework for style-guided image synthesis that seamlessly integrates textual semantic guidance with visual style exemplars, reformulating the task as an in-context style learning problem, supported by analyses of imbalance and noise sensitivity in multimodal attention fusion.
    \item We introduce a dynamic semantic-style integration mechanism that adaptively balances cross-modal attention between text and style features, ensuring high fidelity to both semantic intent and stylistic characteristics.
    \item Extensive experiments demonstrate that our method achieves flexible generalization to unseen styles and textual prompts without requiring model retraining, and can be applied to multiple style-guided image synthesis and various image editing tasks (see Fig.~\ref{fig:p1}).
\end{itemize}

\section{Related Work}
\label{sec:related_work}

\subsection{Personalized Image Generation}
Diffusion models~\cite{wu2023latent,Ramesh:dalle2:2022,Saharia:imagen:2022} have enabled remarkable progress in text-to-image generation by learning to reverse a noise-adding process and synthesizing high-quality images across diverse tasks~\cite{DBLP:conf/cvpr/WangYCWGCLQKW24,Exploiting_Diffusion,FireFlow,multi-turn,Expressive_Image}. Beyond generic generation, personalization has emerged as a key direction, aiming to generate images that preserve user-specific characteristics or follow user-defined visual styles~\cite{zmagic,dreambooth,DreamArtist,UniCanvas}.

Our work focuses on style-guided image synthesis, a sub-task of personalized generation that aims to produce images consistent with the style of a given reference image. Early approaches, such as Textual Inversion~\cite{textual_inversion}, learn a textual embedding from the input image to guide the generation process. DreamBooth~\cite{dreambooth} personalizes a diffusion model by fine-tuning it on a few images of a subject.
The computational demands of DreamBooth's full fine-tuning spurred the development of parameter-efficient methods like Low-Rank Adaptation (LoRA)~\cite{CLoRA,T-LoRA}.
To further improve inference efficiency, some works~\cite{T2I_Adapter,ArtAdapter,StyleAdapter} introduce a lightweight conditioning framework that unlocks precise control over large, frozen text-to-image diffusion models. 

Recently, ReFlow models~\cite{labs2025flux1kontextflowmatching, DBLP:conf/icml/EsserKBEMSLLSBP24, liu:2022:rectifiedflow} adopting diffusion transformer structures have emerged as a promising alternative to diffusion models. This new class of generative models achieves faster inference and enhanced visual fidelity by learning straighter probability paths, as demonstrated in several state-of-the-art text-to-image systems. In parallel, a growing line of research explores in-context personalization in the diffusion transformer model.
In-Context LoRA~\cite{In-Context-LoRA}
introduces a lightweight fine-tuning technique for diffusion transformers. By leveraging multiple images (rather than just text) as context, it efficiently adapts the model using task-specific LoRA modules while keeping the base model weights unchanged, achieving high-fidelity image generation and demonstrating excellent task-agnostic capabilities. Similarly, Diptych Prompting~\cite{Diptych} entirely avoids model fine-tuning by reframing the generation task as a "diptych" inpainting process, enabling precise subject-driven generation in a zero-shot manner.
IP-Prompter~\cite{IP-Prompter} focuses on training-free theme-specific generation, which involves dynamic visual prompts to guide the pre-trained model to recognize and replicate key visual themes from reference images.
 In-Context Brush~\cite{In-Context-Brush} achieves zero-shot subject customizing by reformulating the task within the paradigm of in-context learning.

Despite these advances, effectively integrating semantic content from text with visual style from a reference image in large pretrained diffusion transformers remains challenging. In this work, we explore how content and style can be fused through in-context learning for style-guided text-to-image synthesis, enabling customized stylization with strong content preservation and faithful style alignment.

\subsection{Style Transfer}
Style transfer aims to render the stylistic patterns of a given style image onto a given content image, enabling non-professional users to generate artistic works. With the development of convolutional neural networks (CNNs), Gatys et al.~\cite{gatys:2016:image} discover that perceptual features extracted from pre-trained VGG models can effectively represent both content and style information. Inspired by this work, a series of feed-forward style transfer methods~\cite{Huang:2017:Arbitrary,li:2016:precomputed,li:2017:universal,li:2017:demystifying,park:2019:arbitrary} have been proposed to enable efficient inference. For instance, AdaIN~\cite{Huang:2017:Arbitrary} introduces adaptive instance normalization to merge content and style features, replacing second-order style statistics. WCT~\cite{li:2017:universal} employs whitening and coloring transformations to apply style patterns to content images. SANet~\cite{park:2019:arbitrary} proposes a style-attention mechanism for local content-style fusion. CAST~\cite{zhang:2022:domain} adopts contrastive learning to improve style representation, moving beyond traditional Gram-based losses.
Beyond CNNs, researchers have also explored alternative network architectures for style transfer. ArtFlow~\cite{an:2021:artflow} utilizes a flow-based network to mitigate the content leak problem. StyTr$^2$ ~\cite{Deng:2022:CVPR} designs a pure transformer-based network to better balance content structure and style patterns. More recently, SaMam~\cite{2025cvprsamam} leverages a Mamba-based architecture to achieve a global receptive field while maintaining linear computational complexity.

The advent of diffusion and rectified flow models has further expanded the scope of style transfer. Methods such as InST~\cite{zhang:2023:inversion} and VCT~\cite{cheng:2023:ict} employ inversion-based schemes to encode a style image into a compact style embedding, which guides the generation process. StyleShot~\cite{StyleShot}, InstantStyle~\cite{InstantStyle}, DEADiff~\cite{DEADiff} and CSGO~\cite{CSGO} fine-tune the style-aware adapter module to achieve style transfer. To further reduce computational cost, several training-free approaches have been proposed. For example, Deng et al.~\cite{Deng_2024_CVPR,dengz+} rearrange attention maps in the latent space of pre-trained diffusion models, directly extracting and integrating style priors into content representations without inversion tuning.
WSDT~\cite{yu2025wasserstein} introduces a training-free Wasserstein Style Distribution Transform method for precise and efficient stylized image generation.
VSP~\cite{VSP} preserves content consistency by fusing key style features from the reference in a late self-attention layer.
StyleAligned~\cite{Style_Aligned} proposes a shared attention mechanism to align style and content.

In contrast to existing approaches that depend on fine-tuning or unstable attention replacement, our method introduces a training-free paradigm that leverages style in-context learning for robust and superior content-style integration.


\section{Preliminaries}
\paragraph{Rectified Flow Models}
The fundamental principle of ReFlow models is to learn a straight trajectory between two distributions, such as a noise distribution and a data distribution. This is achieved by training a neural ordinary differential equation (ODE) model that transforms samples from the source distribution to the target distribution along an approximately linear path. Specifically, Rectified Flow constructs a transport map $T$ between two empirical distributions $\pi_0$ (e.g., noise) and $\pi_1$ (e.g., data). It is defined by the ODE:
\begin{equation}
d\mathbf{x}_t = \mathbf{v}_\theta(\mathbf{x}_t, t) \, dt, \quad \mathbf{x}_0 \sim \pi_0, \quad \mathbf{x}_1 \sim \pi_1,
\end{equation}
where $\mathbf{v}_\theta(\mathbf{z}, t)$ denotes the velocity field parameterized by a neural network with parameters $\theta$.

The training objective minimizes the expected squared difference between the predicted velocity and the straight-line direction:
\begin{equation}
\mathcal{L} = \mathbb{E}_{\mathbf{x}_0 \sim \pi_0, \mathbf{x}_1 \sim \pi_1} \left[ \int_0^1 \| \mathbf{x}_1 - \mathbf{x}_0 - \mathbf{v}_\theta(\mathbf{x}_t, t) \|_2^2 \, dt \right],
\end{equation}
where $\mathbf{x}_t = (1-t)\mathbf{x}_0 + t \mathbf{x}_1$.

\paragraph{Diffusion Transformer Structure}
The Diffusion Transformer (DiT) is a novel transformer-based backbone designed for diffusion models, enabling scalable generation and editing across multiple modalities such as text and images~\cite{DBLP:conf/icml/EsserKBEMSLLSBP24}.
DiT consists of a stack of $L$ transformer blocks, each processing concatenated sequences of modality-specific tokens: text tokens $\mathbf{T} \in \mathbb{R}^{N_t \times d}$ (derived from pretrained CLIP~\cite{clip} and T5 models~\cite{2020t5}) and image latents $\mathbf{I} \in \mathbb{R}^{N_i \times d}$ (from the VAE), where $N_t$ and $N_i$ denote sequence lengths, and $d$ is embedding dimensions. 
The forward pass can be formalized as:
\begin{equation}
\mathbf{O}_t, \mathbf{O}_i = \text{DiT}(\mathbf{T}, \mathbf{I}; \theta),
\end{equation}
where $\mathbf{O}_t$ and $\mathbf{O}_i$ are the updated text and image representations, $\theta$ denotes the model parameters. 

The core module in the DiT-based model is the attention mechanism~\cite{BahdanauCB14}, which has become a fundamental component in neural networks for information aggregation:
\begin{equation}
\text{Attn}(\mathbf{Q}, \mathbf{K}, \mathbf{V}) = \text{Softmax}\left(\frac{\mathbf{Q} \mathbf{K}^\top}{\sqrt{d}}\right) \mathbf{V}.
\end{equation}
The attention mechanism serves as the key driver behind these deep token interactions, which are crucial for achieving coherence in generated images.

\section{Method}
\begin{figure*}
\centering
\includegraphics[width=1.0\linewidth]{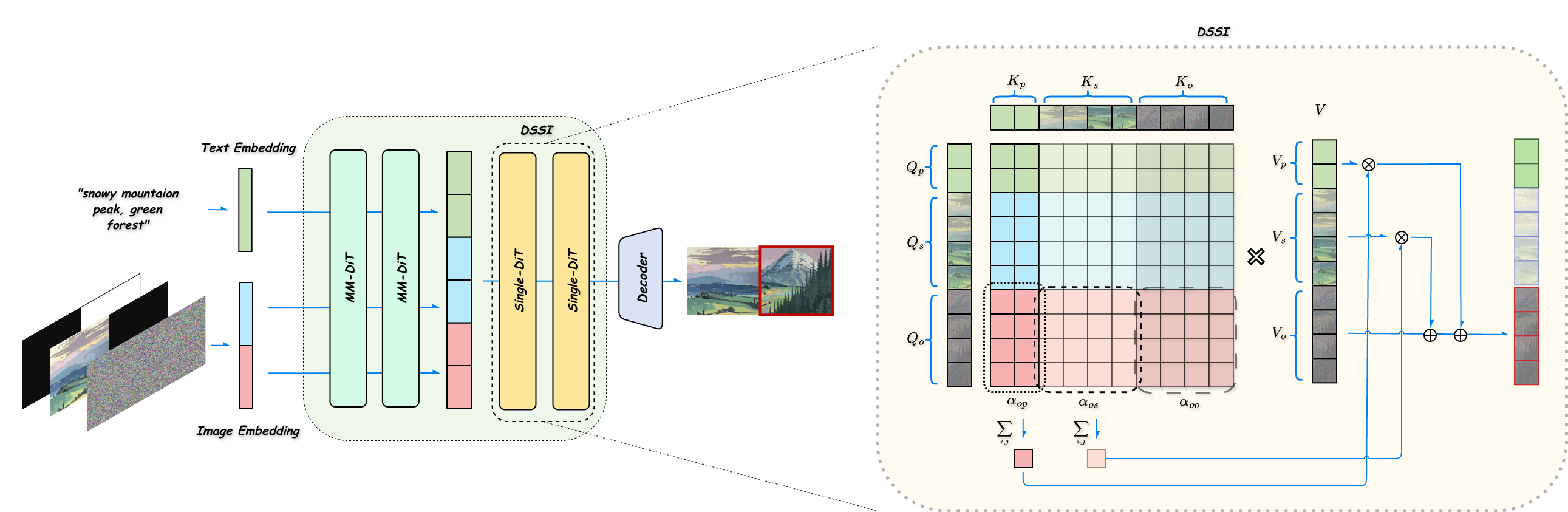}
\caption{Overview of our framework's pipeline. Given a style image $I_s$, we concatenate it with a target mask and input the combined image to the generation network, leveraging its in-context style adaptation capability. The DSSI module adaptively balances attention between textual and visual style cues, yielding outputs that incorporate target style patterns while preserving content semantics.
}
\label{fig:pipline}
\end{figure*}
\subsection{In-context Style Enhancement}
In contrast to the UNet-based text-to-image architecture, which incorporates textual prompts via cross-attention, Transformer-based models integrate both text and image embeddings as a unified token sequence. These multimodal tokens are processed jointly through self-attention within DiT blocks, enabling deeper fusion between semantic and visual representations.

This unified multimodal representation suggests that the model can flexibly attend to multiple types of conditioning information in the same latent space. Building on this insight, we extend the notion of a ``prompt'' beyond text and posit that a style image can serve an equivalent role, acting as a visual conditioning signal that guides the model toward desired stylistic attributes. Instead of relying solely on text to describe abstract visual styles, directly encoding the style image provides explicit visual cues, enabling more accurate and transferable style control.

Given the target style image $\mathbf{I}_s \in \mathbb{R}^{h \times w \times 3}$ and text prompt $\mathbf{c}$, we aim to generate the output image $\mathbf{I}_o \in \mathbb{R}^{h \times w \times 3}$. To leverage the consistency of the DiT-based model, we concatenate $\mathbf{I}_s$ and $\mathbf{I}_o$ as $[\mathbf{I}_s; \mathbf{I}_o]$ and input them as a single image. Since $\mathbf{I}_o$ is unknown, we initialize the input for $\mathbf{I}_o$ to zero and provide a mask $[0; 1]$ to the pre-trained inpainting model. As the generation process occurs in the latent space, we denote the input embedding as $\mathbf{x} = [\mathbf{x}_p; \mathbf{x}_s; \mathbf{x}_o]$, where $\mathbf{x}_p$, $\mathbf{x}_s$, and $\mathbf{x}_o$ represent the text tokens, style image tokens, and output tokens, respectively.

Through the learnable projection matrices $\mathbf{W}_q$, $\mathbf{W}_k$, and $\mathbf{W}_v$, we obtain the query ($\mathbf{Q}$), key ($\mathbf{K}$), and value ($\mathbf{V}$) features in the attention module:
\begin{equation}
\begin{aligned}
\mathbf{Q} &= [\mathbf{Q}_p; \mathbf{Q}_s; \mathbf{Q}_o] = \mathbf{x} \mathbf{W}_q, \\
\mathbf{K} &= [\mathbf{K}_p; \mathbf{K}_s; \mathbf{K}_o] = \mathbf{x} \mathbf{W}_k, \\
\mathbf{V} &= [\mathbf{V}_p; \mathbf{V}_s; \mathbf{V}_o] = \mathbf{x} \mathbf{W}_v.
\end{aligned}
\label{eq:1}
\end{equation}

The attention output can be expressed as:
\begin{equation}
\begin{aligned}
\mathbf{H} &= [\mathbf{H}_p; \mathbf{H}_s; \mathbf{H}_o] = \text{Attn}(\mathbf{Q}, \mathbf{K}, \mathbf{V}) \\
&= \text{Softmax}\left(\frac{\mathbf{Q} \mathbf{K}^\top}{\sqrt{d}}\right) \mathbf{V} \\
&= \text{Softmax}\left(\frac{1}{\sqrt{d}}
\begin{bmatrix}
\mathbf{Q}_p \mathbf{K}_p^\top & \mathbf{Q}_p \mathbf{K}_s^\top & \mathbf{Q}_p \mathbf{K}_o^\top \\
\mathbf{Q}_s \mathbf{K}_p^\top & \mathbf{Q}_s \mathbf{K}_s^\top & \mathbf{Q}_s \mathbf{K}_o^\top \\
\mathbf{Q}_o \mathbf{K}_p^\top & \mathbf{Q}_o \mathbf{K}_s^\top & \mathbf{Q}_o \mathbf{K}_o^\top
\end{bmatrix}\right)
\begin{bmatrix}
\mathbf{V}_p \\ \mathbf{V}_s \\ \mathbf{V}_o
\end{bmatrix}.
\end{aligned}
\end{equation}

For the output corresponding to the query tokens $\mathbf{Q}_o$, the attention can be expressed as:
\begin{equation}
\begin{aligned}
\mathbf{H}_o &= 
\text{Softmax}\left(\frac{1}{\sqrt{d}}
\begin{bmatrix}
\mathbf{Q}_o \mathbf{K}_p^\top & \mathbf{Q}_o \mathbf{K}_s^\top & \mathbf{Q}_o \mathbf{K}_o^\top
\end{bmatrix}\right)
\begin{bmatrix}
\mathbf{V}_p \\ \mathbf{V}_s \\ \mathbf{V}_o
\end{bmatrix} \\
&= \boldsymbol{\alpha}_p \mathbf{V}_p + \boldsymbol{\alpha}_s \mathbf{V}_s + \boldsymbol{\alpha}_o \mathbf{V}_o,
\end{aligned}
\label{fix_atten}
\end{equation}
where $
\boldsymbol{\alpha}_i = 
\frac{\exp\left(\frac{\mathbf{Q}_o \mathbf{K}_i^\top}{\sqrt{d}}\right)}
{\sum_j \exp\left(\frac{\mathbf{Q}_o \mathbf{K}_j^\top}{\sqrt{d}}\right)},
$
with the exponential and softmax operations applied element-wise over the concatenated attention logits for each row of $\mathbf{Q}_o$.

\subsection{Problems in the Vanilla Setting}
\label{sec:vanilla_setting}
In the vanilla formulation, the attention weights $\boldsymbol{\alpha}_p, \boldsymbol{\alpha}_s, \boldsymbol{\alpha}_o$ are derived directly from the concatenated softmax without adaptive reweighting.
This fixed aggregation assumes balanced contributions from the prompt ($\boldsymbol{\alpha}_p \mathbf{V}_p$), style ($\boldsymbol{\alpha}_s \mathbf{V}_s$), and output ($\boldsymbol{\alpha}_o \mathbf{V}_o$) branches, but in practice, it leads to the following issues:

\begin{figure*}
\centering
\includegraphics[width=1.0\linewidth]{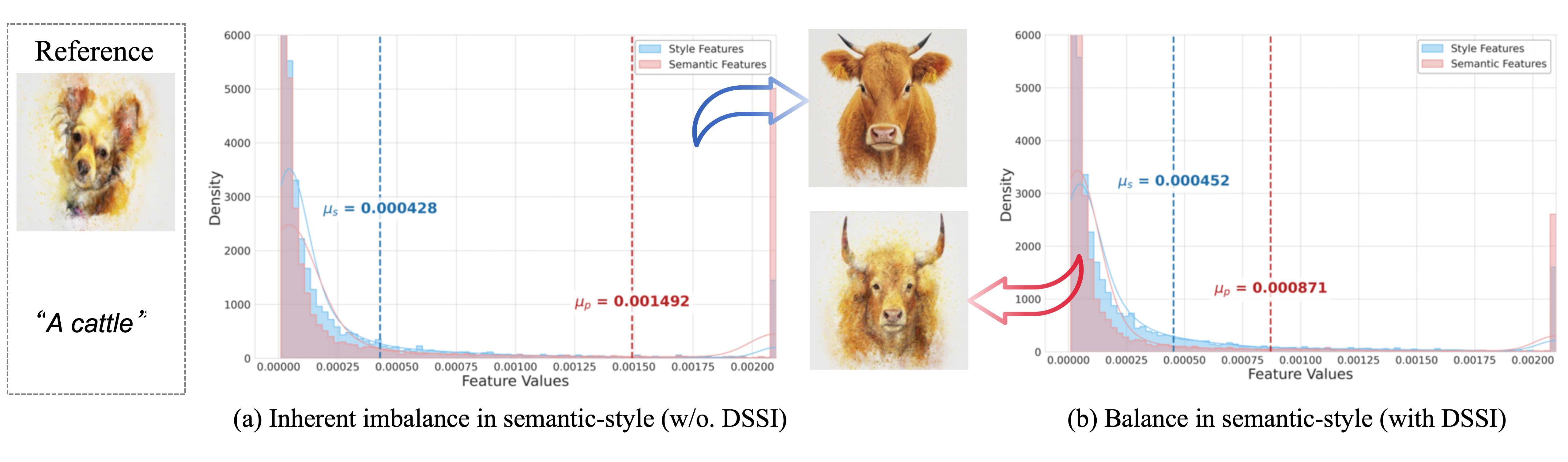}
\caption{An illustration of semantic-style imbalance. Subfigure (a) shows a failure case from the baseline model, where the prompt dominates the output ($\mu_p\gg\mu_s$). In contrast, our approach achieves a more balanced generation where the style ($\mu_s$) and prompt ($\mu_p$) contributions are comparable.}
\label{fig:dssa1_imbalance}
\end{figure*}

\paragraph{Inherent Imbalance in Semantic-Style Contributions}
In the vanilla formulation, the concatenated softmax over the prompt, style, and output branches amplifies differences in activation magnitudes due to the ``winner-takes-all'' behavior of softmax. This normalization exacerbates imbalance when the raw activations from one branch (e.g., style) are systematically larger than another's (e.g., prompt). 

To derive this, recall that the dot products $\vec{\mathbf{q}} \cdot \vec{\mathbf{k}}^T$ for vectors $\vec{\mathbf{q}}, \vec{\mathbf{k}} \in \mathbb{R}^d$ with unit norms have expected variance of 1 after scaling by $1/\sqrt{d}$ (assuming i.i.d. components with variance 1). The softmax then computes probabilities as $\text{Softmax}(\mathbf{Z})_{i,j} = \frac{\exp(\mathbf{Z}_{i,j})}{\sum_k \exp(\mathbf{Z}_{i,k})}$, where $\mathbf{Z} = \frac{1}{\sqrt{d}} [\mathbf{Q}_o \mathbf{K}_p^\top \quad \mathbf{Q}_o \mathbf{K}_s^\top \quad \mathbf{Q}_o \mathbf{K}_o^\top]$ (concatenated across branches). For activations with unit variance, softmax exhibits a strong winner-takes-all effect: if one branch's mean activation value $\mu_s$ (for style branch $\vec{\mathbf{q}_o} \mathbf{K}_s$) exceeds another's $\mu_p$ (for prompt branch $\vec{\mathbf{q}_o} \mathbf{K}_p$) by a large margin (e.g., $\mu_s - \mu_p > 1$), the probability mass concentrates on the higher branch, as $\exp(\mu_s) \gg \exp(\mu_p)$. Quantitatively, let the concatenated activations have branch means $\mu_p, \mu_s, \mu_o$, we have the following statement.
\begin{proposition}
\label{proposition_1}
Assume that for each branch $b \in \{p,s,o\}$, the entries of $\mathbf{Z}$ satisfy $\mathbf{Z}_{i,j} = \mu_b + \xi_{i,j}$,
where $\mu_b$ is the branch-specific mean and $\xi_{i,j}$ are i.i.d.\ zero-mean random variables with variance $\sigma^2 \approx 1$. Then the $\ell_1$ norm of each row of the style-branch attention weights satisfies:
$$\|\vec{\alpha}_s\|_1 \approx \frac{N_s \exp(\mu_s)}{N_p \exp(\mu_p) + N_s \exp(\mu_s) + N_o \exp(\mu_o)} = \frac{1}{1 + \frac{N_p}{N_s} \exp(\mu_p - \mu_s) + \frac{N_o}{N_s} \exp(\mu_o - \mu_s)},$$ up to an approximation error of order $\mathcal{O}(\sigma^3)$.
\end{proposition}
The implications are immediate: when $\mu_s \gg \mu_p$, the attention becomes dominated by the style branch, driving $\|\vec{\alpha}_s\|_1 \to 1$ and $\|\vec{\alpha}_p\|_1 \to 0$, which produces over-stylization and semantic distortion. Conversely, as shown in Fig. \ref{fig:dssa1_imbalance}, when $\mu_p \gg \mu_s$, the style branch is effectively suppressed, giving weak or washed-out stylization. Although the $1/\sqrt{d}$ scaling ensures variance stability as the dimension grows, it does not prevent inappropriate cross-branch imbalances: differences arising from input statistics (e.g., texture-heavy style encodings vs.\ semantic prompt encodings) can push the system into regimes where one branch almost fully dominates.

\begin{figure}
    \centering
    \begin{subfigure}[b]{0.31\textwidth}
        \centering
        \includegraphics[width=\linewidth]{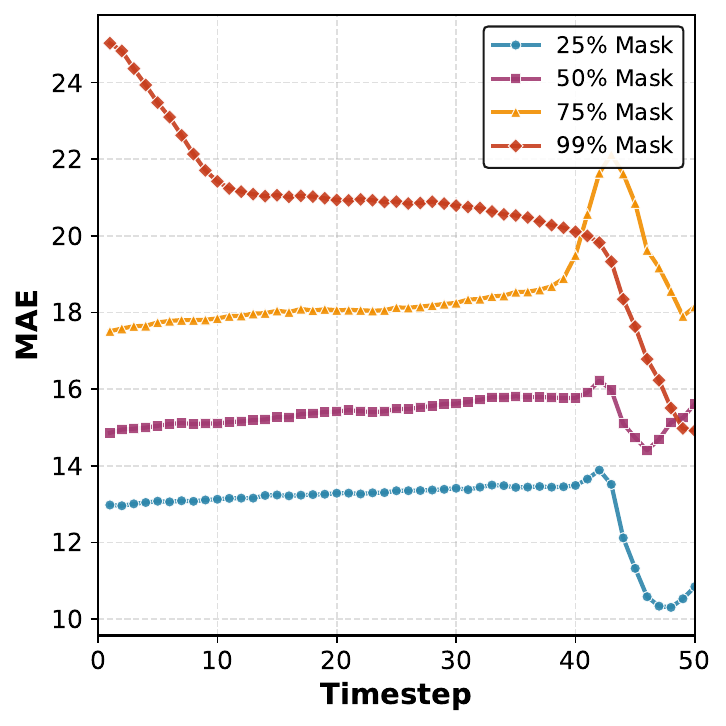}
        \subcaption{MAE of Softmax logits $\tilde{\mathbf{Z}}$ v.s. $\mathbf{Z}$.}
        \label{fig:logits}
    \end{subfigure}
    \hfill
    \begin{subfigure}[b]{0.31\textwidth}
        \centering
        \includegraphics[width=\linewidth]{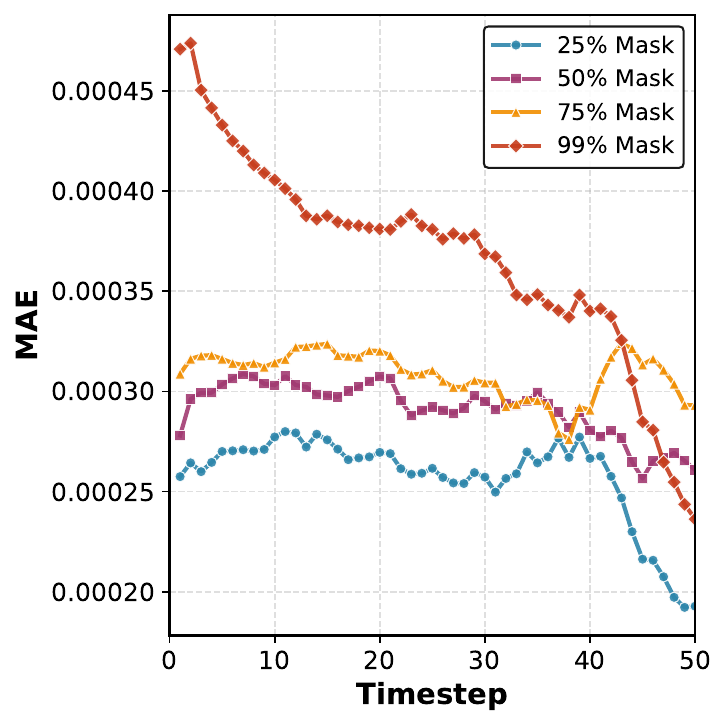}
        \subcaption{MAE of attention weights $\boldsymbol{\tilde{\alpha}}$ v.s. $\boldsymbol{\alpha}$.}
        \label{fig:attention_weights}
    \end{subfigure}
     \hfill
    \begin{subfigure}[b]{0.31\textwidth}
        \centering
        \includegraphics[width=\linewidth]{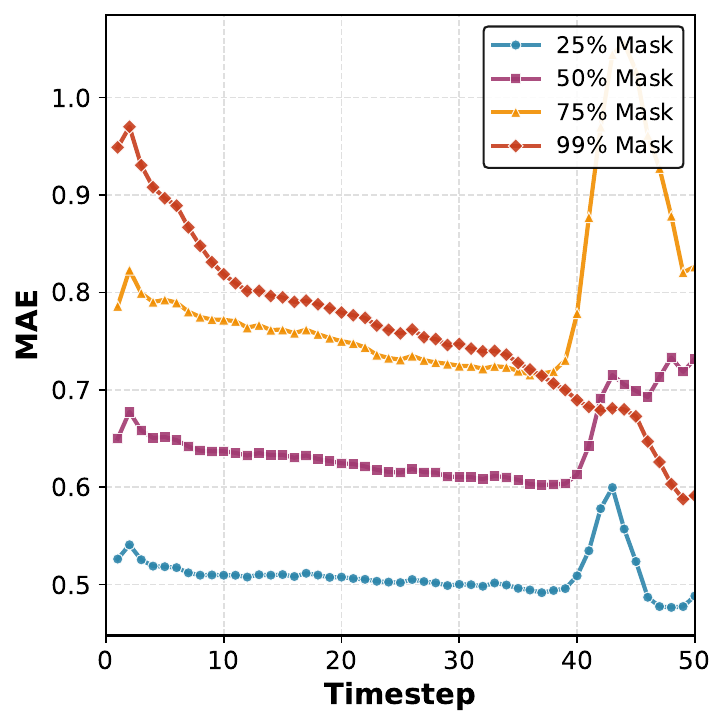}
        \subcaption{MAE of output tensor $\tilde{\mathbf{h}}$ v.s. $\mathbf{h}$.}
        \label{fig:attention_weights}
    \end{subfigure}
    \caption{An illustration depicting how noise introduced by masking the input propagates through logits, attention weights, and the output tensor. The Mean Absolute Error (MAE) quantifies the differences between results obtained with masking and the reference results. As the masked area increases, these differences become more pronounced.}
    \label{fig:noise_sensitivity}
\end{figure}

\paragraph{Sensitivity to Noise}
In the previous subsection, we demonstrated that the vanilla attention configuration becomes unsatisfactory when the model strongly favors either the content branch ($\mu_p \gg \mu_s$) or the style branch ($\mu_s \gg \mu_p$). A less obvious, yet equally significant issue arises from perturbations introduced by the input mask in the inpainting setting. 
To investigate this, we conduct a zero-mask reconstruction experiment: within the FLUX-Fill pipeline, we concatenate a reference image with the same image masked at different percentages ($25\%$, $50\%$, $75\%$ and $99\%$) and attempt to reconstruct the original.
As shown in Fig. \ref{fig:noise_sensitivity}, applying a zero-mask for reconstruction perturbs the activations in the attention operations relative to those produced by a clean input. These perturbations are further amplified by the softmax operation, ultimately causing the output to diverge from the original image. In our style-guide image synthesis experiments, we may also regard image synthesis guided by a prompt as the reconstruction of an oracle image. Therefore, reducing the noise introduced by the zero-mask is essential for achieving higher-quality generation. 

Formally, let
$
\tilde{\mathbf{Z}} = \mathbf{Z} + \boldsymbol{\epsilon}$, perturbed by $ \|\boldsymbol{\epsilon}\|_\infty \le \delta,
$
and $\tilde{\boldsymbol{\alpha}}_* = \operatorname{Softmax}(\tilde{\mathbf{Z}}_*)$. The following result characterizes how logit perturbations propagate through the softmax.

\begin{proposition}
Let ${s} \in \mathbb{R}^n$ be an arbitrary logit vector and let ${\alpha} = \operatorname{Softmax}({s})$ denote the corresponding probability distribution, i.e.,
$\alpha_j = \frac{\exp(s_j)}{\sum_{k=1}^n \exp(s_k)}, \quad j = 1,\dots,n.$
Let $\tilde{{s}} \in \mathbb{R}^n$ be a perturbed logit vector satisfying
$\|\tilde{{s}} - {s}\|_\infty \leq \delta$
for some $\delta \geq 0$, and define $\tilde{{\alpha}} = \operatorname{Softmax}(\tilde{{s}})$.
Then the total variation distance between the two distributions is bounded by
$\operatorname{Total\_Variance}(\tilde{{\alpha}}, {\alpha}) := \frac{1}{2} \|\tilde{{\alpha}} - {\alpha}\|_1 \leq 1 - e^{-2\delta}.$
Equivalently,
$\|\tilde{{\alpha}} - {\alpha}\|_1 \leq 2(1 - e^{-2\delta}).$
For small perturbations $\delta$, the bound behaves asymptotically as
$\|\tilde{{\alpha}} - {\alpha}\|_1 \leq 4\delta + O(\delta^2).$
\label{proposition_2}
\end{proposition}

Using this bound, the perturbation of the output row $\tilde{\mathbf{h}}_o$ compared to the ideal case $\mathbf{h}_o$ satisfies:
\begin{equation}
\begin{aligned}
\|\tilde{\mathbf{h}}_o - \mathbf{h}_o\|_2
&= \|(\vec{\tilde{\alpha}}_p - \vec{\alpha}_p)\mathbf{V}_p
    + (\vec{\tilde{\alpha}}_s - \vec{\alpha}_s)\mathbf{V}_s
    + (\vec{\tilde{\alpha}}_o - \vec{\alpha}_o)\mathbf{V}_o\|_2 \\
&\le \sum_* \|\vec{\tilde{\alpha}}_* - \vec{\alpha}_*\|_1\,\|\mathbf{V}_*\|_2 \\
&\le \max_*\left(\|\vec{\tilde{\alpha}}_* - \vec{\alpha}_*\|_1\right)\,
    (\|\mathbf{V}_p\|_2 + \|\mathbf{V}_s\|_2 + \|\mathbf{V}_o\|_2)\\
&= 2(1 - e^{-2\delta})\,
    (\|\mathbf{V}_p\|_2 + \|\mathbf{V}_s\|_2 + \|\mathbf{V}_o\|_2),
\end{aligned}
\label{noisy_output}
\end{equation}
where $\alpha_*$ denotes either $\alpha_p$, $\alpha_s$ or $\alpha_o$. The factor $2(1 - e^{-2\delta})$ is intrinsic, reflecting the stochastic noise inherent to the FLUX-Fill pipeline. Therefore, we focus on the simplest strategy: reducing $\max_*\left(\|\vec{\tilde{\alpha}}_* - \vec{\alpha}_*\|_1\right)$, which in turn leads to a tighter upper bound on the output perturbation.

\subsection{Dynamic Semantic-Style Integration}

In the above formulation, the attention output \(\mathbf{h}_o\) integrates three types of contextual information: semantic guidance from textual prompts (\(\mathbf{Q}_o, \mathbf{K}_p, \mathbf{V}_p\)), stylistic cues from the reference image (\(\mathbf{Q}_o, \mathbf{K}_s, \mathbf{V}_s\)), and output (\(\mathbf{Q}_o, \mathbf{K}_o, \mathbf{V}_o\)).  
Ideally, these sources should complement each other, text guiding the global semantic structure, while the style image provides local texture and color cues. However, in practice, we observe that the contributions of semantic and style signals are often unbalanced. Specifically, when the style image exhibits strong visual features, its attention response tends to dominate the update of \(\mathbf{h}_o\), leading to over-stylization, content distortion, or unintended introduction of style structures. Conversely, when the text prompt exerts the primary influence, the model may overemphasize textual semantics, resulting in weak style transfer and diminished visual fidelity.  
This imbalance indicates that a fixed attention formulation (as in Eq.~\eqref{fix_atten}) cannot dynamically adapt to varying semantic-style relationships during inference.

To address this, we introduce a Dynamic Semantic-Style Integration (DSSI) mechanism, which adaptively adjusts the attention balance between textual and visual style cues, enabling the model to maintain semantic consistency while preserving fine-grained stylistic fidelity. This approach draws inspiration from gated fusion mechanisms in multimodal vision-language models, where adaptive weights prevent one modality from overwhelming others, ensuring coherent integration without training.

We reformulate DSSI as an optimization problem over the attention outputs. Define the semantic-guided reference output as:
\begin{equation}
    \mathbf{s} = \kappa\cdot\boldsymbol{\alpha}_p \mathbf{V}_p + \boldsymbol{\alpha}_o \mathbf{V}_o,
\end{equation}
emphasizing text prompts and output, ignoring style, and the style-guided reference output as:
\begin{equation}
    \mathbf{t} = \kappa\cdot\boldsymbol{\alpha}_s \mathbf{V}_s + \boldsymbol{\alpha}_o \mathbf{V}_o,
\label{tab:kappa}
\end{equation}
emphasizing the style reference and output, ignoring text prompts, where $\kappa$ serves as a hyper-parameter to adjust the strength of the prior of prompt and style. The goal is to find a fused output:
\begin{equation}
    \mathbf{h}_o^{\text{DSSI}} = (1 - \lambda) \mathbf{s} + \lambda \mathbf{t},
\end{equation}
parameterized by \(\lambda \in [0,1]\), that balances proximity to \(\mathbf{s}\) (for semantic fidelity) and \(\mathbf{t}\) (for stylistic fidelity). This can be formalized as the following unconstrained optimization problem:
\begin{equation}
\min_{\lambda \in [0,1]} \, \mathcal{L}(\lambda) = \| \mathbf{h}(\lambda) - \mathbf{s} \|_F^2 + \gamma \| \mathbf{h}(\lambda) - \mathbf{t} \|_F^2,
\end{equation}
where \(\mathbf{h}(\lambda) = (1 - \lambda) \mathbf{s} + \lambda \mathbf{t}\), \(\| \cdot \|_F^2\) denotes the squared Frobenius norm, and \(\gamma > 0\) is a trade-off parameter controlling the relative emphasis on stylistic versus semantic fidelity. The Frobenius norm is particularly apt here as it respects the matrix structure of attention outputs (rows as tokens, columns as dimensions), providing a holistic measure of discrepancy while being computationally efficient and differentiable.

Since \(\mathbf{h}(\lambda)\) is linear in \(\lambda\), \(\mathcal{L}(\lambda)\) is a convex quadratic function, admitting a closed-form solution. To confirm convexity, note the second derivative: \(\frac{d^2}{d\lambda^2} \mathcal{L}(\lambda) = 2D + 2\gamma D = 2D(1 + \gamma) > 0\) (for \(D > 0\)), ensuring a unique global minimum and stable optimization during inference. Let \(D = \| \mathbf{t} - \mathbf{s} \|_F^2\) (assuming \(D > 0\); otherwise, \(\mathbf{s} = \mathbf{t}\) and any \(\lambda\) suffices). Substituting \(\mathbf{h}(\lambda)\) yields:
\begin{equation}
\begin{aligned}
\| \mathbf{h}(\lambda) - \mathbf{s} \|_F^2 &= \| \lambda (\mathbf{t} - \mathbf{s}) \|_F^2 = \lambda^2 D, \\
\| \mathbf{h}(\lambda) - \mathbf{t} \|_F^2 &= \| (1 - \lambda) (\mathbf{s} - \mathbf{t}) \|_F^2 = (1 - \lambda)^2 D.
\end{aligned}
\end{equation}
Thus,
\begin{equation}
\begin{aligned}
\mathcal{L}(\lambda) = D \left[ \lambda^2 + \gamma (1 - \lambda)^2 \right].
\end{aligned}
\end{equation}
Minimizing with respect to \(\lambda\) (ignoring the constant \(D\)):
\begin{equation}
\begin{aligned}
&\frac{d}{d\lambda} \left[ \lambda^2 + \gamma (1 - \lambda)^2 \right] 
= 2\lambda - 2\gamma (1 - \lambda) = 0 \\
&\implies \lambda + \gamma \lambda = \gamma \\
&\implies \lambda^* = \frac{\gamma}{1 + \gamma}.
\end{aligned}
\end{equation}
Since \(\gamma \geq 0\), \(\lambda^* \in [0,1]\), satisfying the bounds without additional constraints. Alternatively, this can be interpreted probabilistically: treating \(\mathbf{s}\) and \(\mathbf{t}\) as ``priors'' from semantic and style distributions, \(\lambda^*\) acts as a posterior weight, akin to Bayesian model averaging, where \(\gamma\) reflects the relative ``evidence'' from each modality.

The trade-off parameter \(\gamma\) is intuitively selected as the ratio of style integration strength to semantic integration strength, i.e., \(\gamma = \frac{\lambda_p}{\lambda_s}\). Here, \(\lambda_p\) and \(\lambda_s\) measure the overall softmax alignment between the target queries and the respective keys:
\begin{equation}
\begin{aligned}
\lambda_p &= \log \sum_{i=1}^{N_o} \sum_{j=1}^{N_p} \left[ \text{Softmax}\left(\frac{\mathbf{Q}_o \mathbf{K}_p^\top}{\sqrt{d}}\right) \right]_{i,j}, \\
\lambda_s &= \log \sum_{i=1}^{N_o} \sum_{j=1}^{N_s} \left[ \text{Softmax}\left(\frac{\mathbf{Q}_o \mathbf{K}_s^\top}{\sqrt{d}}\right) \right]_{i,j},
\end{aligned}
\end{equation}
where \(N_o\), \(N_p\), and \(N_s\) are the number of target, prompt, and style tokens, respectively. This choice ensures that if the style cues are strongly aligned (\(\lambda_s \gg \lambda_p\)), \(\gamma\) is small, prioritizing semantic preservation (\(\lambda^* \approx 0\)). Conversely, if semantic cues dominate (\(\lambda_p \gg \lambda_s\)), \(\gamma\) is small, favoring stylistic fidelity (\(\lambda^* \approx 1\)).

Substituting \(\gamma = \frac{\lambda_p}{\lambda_s}\) into the closed-form solution derives our DSSI weighting:
\begin{equation}
\lambda = \frac{\gamma}{1 + \gamma} = \frac{\frac{\lambda_p}{\lambda_s}}{1 + \frac{\lambda_p}{\lambda_s}} = \frac{\lambda_p}{\lambda_s + \lambda_p}.
\end{equation}
The reweighted output attention is then:
\begin{equation}
\mathbf{h}_o^{\text{DSSI}} = \kappa\cdot\big((1 - \lambda)\cdot\boldsymbol{\alpha}_p \mathbf{V}_p + \lambda\cdot \boldsymbol{\alpha}_s \mathbf{V}_s \big)+ \boldsymbol{\alpha}_o \mathbf{V}_o.
\label{eq:dssa}
\end{equation}
This mechanism allows the model to adaptively modulate the contribution of textual and visual cues based on the current attention distribution, effectively mitigating conflicts between semantic and stylistic features to make a balance. When the textual prompt provides strong semantic cues, $\lambda$ increases to enhance stylistic fidelity. Conversely, when the style reference exhibits rich texture patterns, the model preserves more structural content by reducing $\lambda$.

\begin{figure}
    \centering
    \begin{subfigure}[b]{0.48\textwidth}
        \centering
        \includegraphics[width=\linewidth]{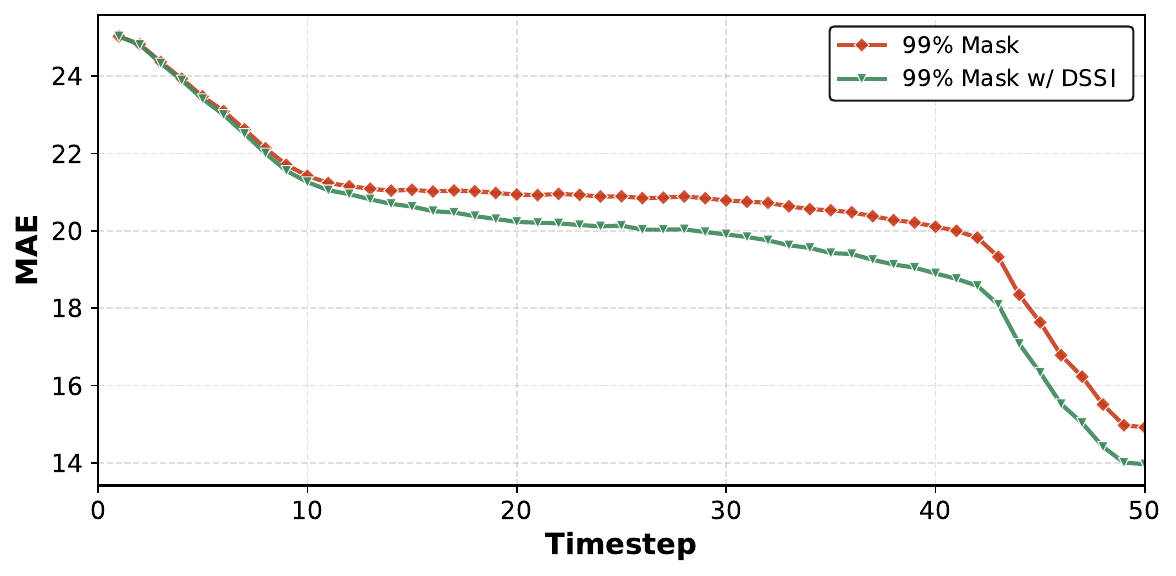}
        \subcaption{MAE of Softmax logits $\tilde{\mathbf{Z}}$ v.s. $\mathbf{Z}$.}
        \label{fig:logits}
    \end{subfigure}
    \hfill
    \begin{subfigure}[b]{0.48\textwidth}
        \centering
        \includegraphics[width=\linewidth]{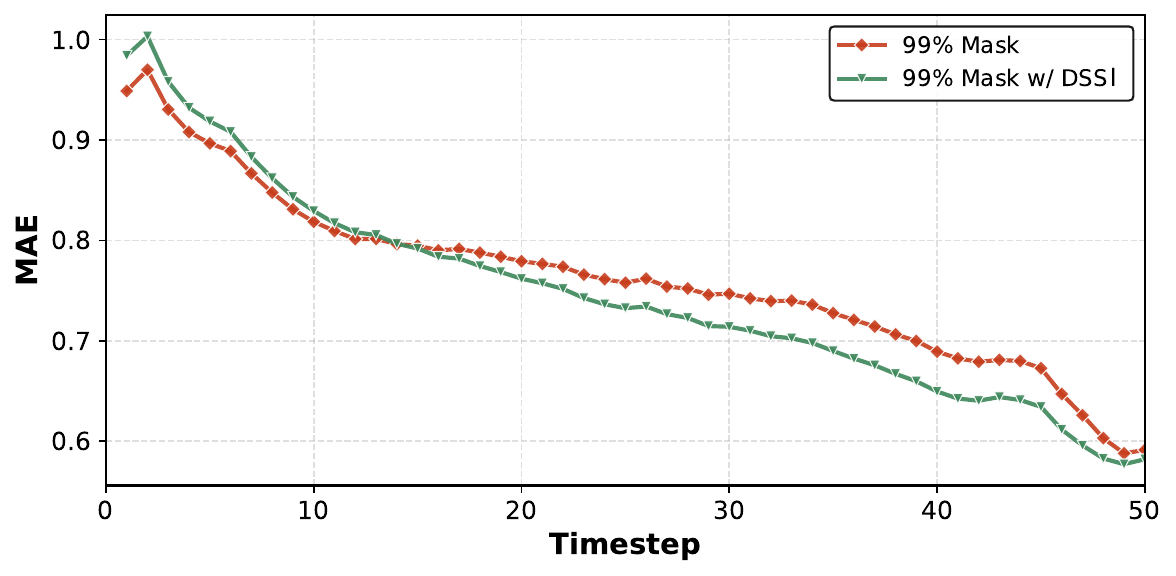}
        \subcaption{MAE of output tensor $\tilde{\mathbf{h}}$ v.s. $\mathbf{h}$.}
        \label{fig:attention_weights}
    \end{subfigure}
    \caption{An illustration showing the effect of DSSI in reducing the noise introduced by the zero-mask, with the reference image serving as the baseline for MAE computation.}
    \label{fig:noise_sensitivity_ours}
\end{figure}

\subsection{Noise Sensitivity Analysis}
Having shown that the proposed method encourages more balanced attention between the prompt and style branches, we now analyze how the scaling factor $\lambda$ contributes to robustness under noise. Let the clean DSSI output be
$
\mathbf{H}_o 
= (1 - \lambda)\,\boldsymbol{\alpha}_p \mathbf{V}_p 
  + \lambda\,\boldsymbol{\alpha}_s \mathbf{V}_s 
  + \boldsymbol{\alpha}_o \mathbf{V}_o,
$
where we set $\kappa = 1$ for simplicity, 
$\lambda = \frac{\lambda_p}{\lambda_p + \lambda_s}$ is the dynamic scaling coefficient, 
and $\boldsymbol{\alpha}_p, \boldsymbol{\alpha}_s, \boldsymbol{\alpha}_o$ are the attention weights produced by the concatenated softmax.  
Under the noise introduced by our training-free modification for the pre-trained FLUX-Fill model, the output becomes
$
\tilde{\mathbf{H}}_o
= (1 - \tilde{\lambda})\,\tilde{\boldsymbol{\alpha}}_p \mathbf{V}_p
  + \tilde{\lambda}\,\tilde{\boldsymbol{\alpha}}_s \mathbf{V}_s
  + \tilde{\boldsymbol{\alpha}}_o \mathbf{V}_o,
$
with $\tilde{\lambda} = \frac{\tilde{\lambda}_p}{\tilde{\lambda}_p + \tilde{\lambda}_s}$ and perturbed attention weights 
$\tilde{\boldsymbol{\alpha}}_*$.  
Subtracting the clean output yields:
\begin{equation}
\begin{aligned}
\tilde{\mathbf{H}}_o - \mathbf{H}_o
&= (1 - \tilde{\lambda}) \tilde{\boldsymbol{\alpha}}_p \mathbf{V}_p 
   - (1 - \lambda)\boldsymbol{\alpha}_p \mathbf{V}_p
   + \tilde{\lambda}\,\tilde{\boldsymbol{\alpha}}_s \mathbf{V}_s 
   - \lambda\boldsymbol{\alpha}_s \mathbf{V}_s
   + (\tilde{\boldsymbol{\alpha}}_o - \boldsymbol{\alpha}_o)\mathbf{V}_o \\
&= (1 - \lambda)(\tilde{\boldsymbol{\alpha}}_p - \boldsymbol{\alpha}_p)\mathbf{V}_p
   - (\tilde{\lambda} - \lambda)\tilde{\boldsymbol{\alpha}}_p \mathbf{V}_p + \lambda(\tilde{\boldsymbol{\alpha}}_s - \boldsymbol{\alpha}_s)\mathbf{V}_s\\
   & \qquad 
   + (\tilde{\lambda} - \lambda)\tilde{\boldsymbol{\alpha}}_s \mathbf{V}_s
   + (\tilde{\boldsymbol{\alpha}}_o - \boldsymbol{\alpha}_o)\mathbf{V}_o.
\end{aligned}
\end{equation}
Applying the triangle inequality row-wise and using the same norm relations as in Eq.~\eqref{noisy_output}, we obtain:
\begin{equation}
\begin{aligned}
        \|\tilde{\mathbf{h}}_o - \mathbf{h}_o\|_2 \leq &\underbrace{(1 - \lambda) \|\vec{\tilde{{\alpha}}}_p - \vec{\alpha}_p\|_1 \|\mathbf{V}_p\|_2+\lambda \|\vec{\tilde{\alpha}}_s - \vec{\alpha}_s\|_1 \|\mathbf{V}_s\|_2}_{\text{term I}}\\
        & + \underbrace{|\tilde{\lambda} - \lambda| \|\vec{\tilde{{\alpha}}}_p\|_2 \|\mathbf{V}_p\|_2 + |\tilde{\lambda} - \lambda|\|\vec{\tilde{{\alpha}}}_s\|_2 \|\mathbf{V}_s\|_2}_{\text{term II}} \\
        & + \underbrace{\|\vec{\tilde{{\alpha}}}_o - \vec{\alpha}_o\|_2 \|\mathbf{V}_o\|_2}_{\text{term III}}.
\end{aligned}
\label{dssa_bound}
\end{equation}
Term~III matches the vanilla setting exactly.  
Term~I is always smaller than its vanilla counterpart because $\lambda \in [0,1]$. We now show that Term~II is negligible.

\begin{proposition}
Let $\lambda_p = \log \sum_{i,j} [\operatorname{Softmax}(\mathbf{Q}_o \mathbf{K}_p^\top / \sqrt{d})]_{i,j}$  
and $\lambda_s = \log \sum_{i,j} [\operatorname{Softmax}(\mathbf{Q}_o \mathbf{K}_s^\top / \sqrt{d})]_{i,j}$  
denote the clean alignment strengths for the prompt and style branches, respectively, and 
$\lambda = \frac{\lambda_p}{\lambda_p + \lambda_s}$.  
Under noisy alignments  
$\tilde{\lambda}_p = \lambda_p + \Delta_p$ and $\tilde{\lambda}_s = \lambda_s + \Delta_s$  
with noise bounds $|\Delta_p| \le \epsilon_p$ and $|\Delta_s| \le \epsilon_s$,  
the perturbed scaling factor 
$\tilde{\lambda} = \frac{\tilde{\lambda}_p}{\tilde{\lambda}_p + \tilde{\lambda}_s}$
satisfies
$
|\tilde{\lambda} - \lambda|
\le \frac{\max(\epsilon_p,\epsilon_s)}{\lambda_p + \lambda_s}.
$
\end{proposition}

Following the noise model in Section~\ref{sec:vanilla_setting},
$
\tilde{\mathbf{Z}} = \mathbf{Z} + \boldsymbol{\epsilon},  \|\boldsymbol{\epsilon}\|_\infty \le \delta,
$
we further obtain $\max(\epsilon_p,\epsilon_s)\le 2\delta$ (see Appendix).  
Since $\|\vec{\tilde{\alpha}}_*\|_1 \le 1$, Term~II satisfies:
\[
|\tilde{\lambda} - \lambda|\, \|\vec{\tilde{\alpha}}_*\|_2 
\le |\tilde{\lambda} - \lambda| \|\vec{\tilde{\alpha}}_*\|_1
\le \frac{2\delta}{\lambda_p + \lambda_s}.
\]
In both the imbalance regime (where one branch dominates, e.g., $\lambda_p\gg\lambda_s$) and the balanced regime (where prompt and style have comparable strength, e.g., $\lambda_p\approx\lambda_s$), we always have $\lambda_p + \lambda_s \gg \delta$.  
Thus, Term~II is vanishingly small and can be ignored. Therefore, Eq. \eqref{dssa_bound} clearly shows a lower bound than Eq. \eqref{noisy_output}; hence, the proposed method is less sensitive to the noise introduced by the mask for stylized image generation than the vanilla setting. To further illustrate this, we present a comparison of the effect of DSSI on noise reduction during masked image reconstruction in Fig.~\ref{fig:noise_sensitivity_ours}. The results demonstrate that DSSI progressively reduces noise in the logits and diminishes its impact on the outputs.

\section{Experiment}

\subsection{Evaluation}
\begin{figure*}
\centering
\includegraphics[width=1.0\linewidth]{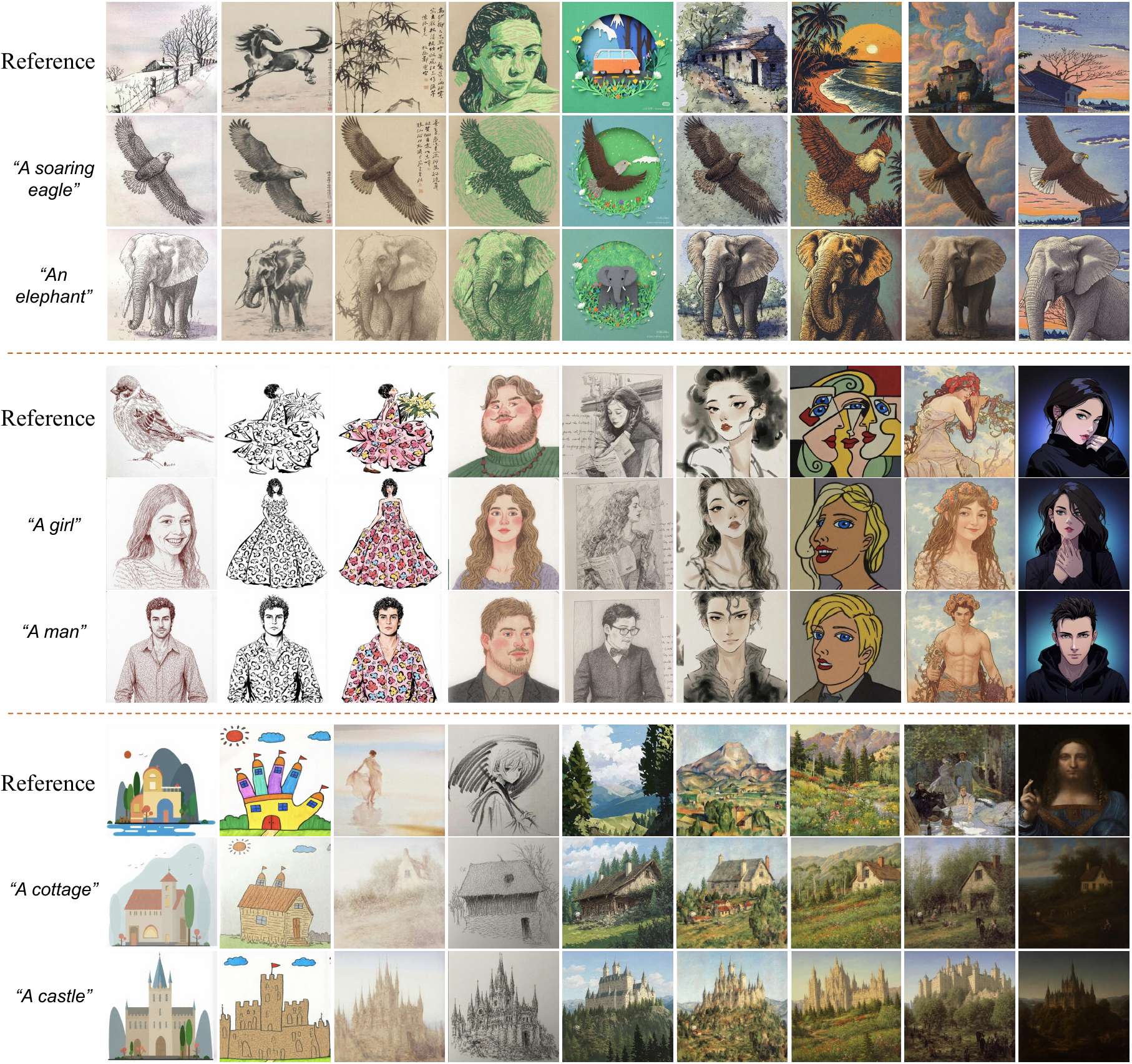}
\caption{Demonstration of our method's generation results across diverse style references and content scenarios, showcasing robust stylization while preserving semantic fidelity.
}
\label{fig:experiment_our}
\end{figure*}

Our method uses Flux-Fill~\cite{blackforestlabs2023flux} as the base generation model. All experiments are conducted on a single NVIDIA Tesla P40 GPU. Unless otherwise specified, we use the default hyperparameters provided by the original Flux-Fill repository and perform inference with 30 denoising steps.
As demonstrated in Fig.~\ref{fig:experiment_our}, our approach exhibits remarkable generalization capability across a wide variety of style references and content scenarios.
In addition, we comprehensively compare our approach against several state-of-the-art style-guided image synthesis methods, including WSDT~\cite{yu2025wasserstein}, VSP~\cite{VSP}, StyleShot~\cite{StyleShot}, StyleAligned~\cite{Style_Aligned}, DEADiff~\cite{DEADiff}, InstaStyle~\cite{INSTASTYLE}, InstantStyle~\cite{InstantStyle} and CSGO~\cite{CSGO}.
The generation process for a 512 × 512 image takes approximately 15 seconds. This inference speed is on par with training-free baselines (e.g., WSDT, VSP, StyleAligned) and significantly faster than methods requiring day-level training (e.g., StyleShot, DEADiff, InstaStyle, InstantStyle, CSGO).

\subsubsection{Qualitative Evaluation}
\begin{figure*}
\centering
\includegraphics[width=1.0\linewidth]{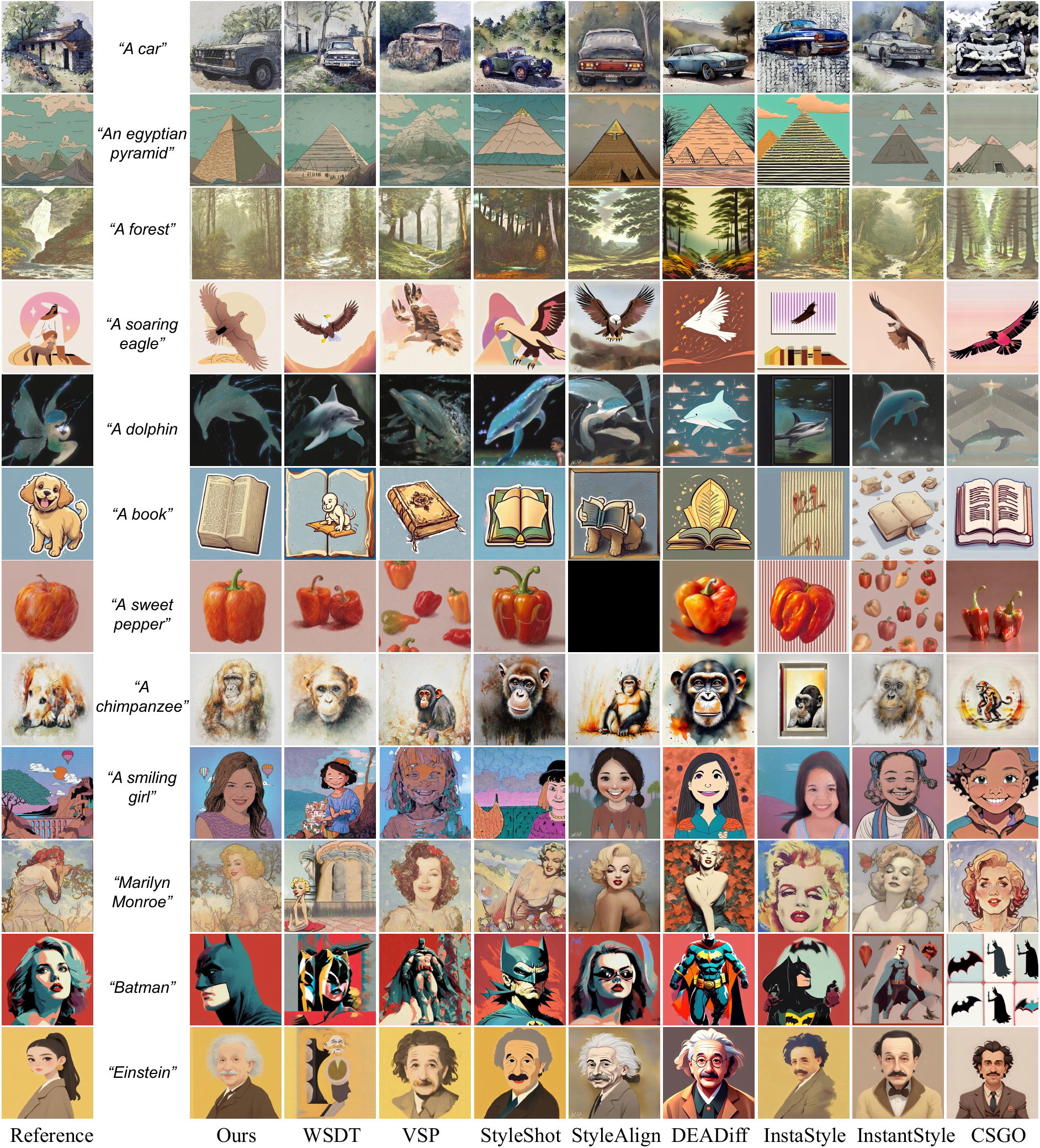}
\caption{Comparison with state-of-the-art methods. 
}
\label{fig:experiment}
\end{figure*}

As shown in Fig.~\ref{fig:experiment}, we present representative qualitative comparisons with SOTA baselines on a diverse set of challenging content prompts and reference style images.
Existing methods exhibit clear limitations in achieving robust style consistency and content fidelity simultaneously.
StyleAligned, DEADiff, InstantStyle, InstaStyle and CSGO frequently fail to faithfully reproduce the target style, resulting in outputs that either marginally reflect the reference (weak stylization) or introduce inconsistent and patchy style patterns.
StyleShot, VSP and WSDT demonstrate improved style expressiveness in relatively simple cases. However, when confronted with complex cases, these methods struggle to effectively adapt the style elements to the content layout, such as the ``A smiling girl'' and ``A dolphin'' cases.
As can be observed, our approach achieves superior visual fidelity in both style expressiveness and content preservation, producing stylized images that closely adhere to the reference style while maintaining the semantic structure of the input content. Importantly, beyond mere low-level attributes such as color distribution and local texture, our approach effectively learns and transfers higher-order stylistic features, including global composition principles, rhythm of brushstrokes, spatial organization of visual elements, and overall artistic intent, as illustrated in the fourth and sixth rows of Fig.~\ref{fig:experiment}.

\subsubsection{Quantitative Evaluation}
This section presents a comprehensive quantitative comparison between the proposed method and existing state-of-the-art approaches to validate its effectiveness.

\textbf{Evaluation Metrics.} 
Following InstaStyle~\cite{INSTASTYLE}, we use the object classes from CIFAR-100~\cite{Krizhevsky2009LearningML} and their corresponding class names as text prompts to generate content images. For style references, we adopt the original 60 artistic style images curated by InstaStyle and further enrich the dataset with 40 additional diverse style images (including abstract, sketch painting, cartoon, etc.), resulting in a total of 100 style references. These additional images are provided in the supplementary material.
Performance is measured from two complementary perspectives using a pretrained CLIP~\cite{clip}.
\textbf{Content Alignment}: Cosine similarity between the CLIP image embedding of the generated result and the CLIP text embedding of the input content prompt (higher is better).
\textbf{Style Alignment}: Cosine similarity between the CLIP image embedding of the generated result and the CLIP image embedding of the reference style image (higher is better).

As summarized in Table~\ref{tab:quancomps}(a), our method achieves the highest style consistency score while maintaining competitive content alignment. This result confirms our approach's unique capability to strike an optimal balance, producing images that exhibit both strong fidelity to the content prompt and faithful adherence to the reference style.

\textbf{User Study.} To further evaluate subjective quality, we conducted a user preference study involving 30 participants, evenly split between technical ($50\%$) and non-technical ($50\%$) backgrounds. Each participant was asked to rate the generated results on three criteria (including content preservation, style alignment, and overall visual quality) on a scale of 1-5 (1 = poorest, 5 = best). The evaluation included 30 randomly selected cases generated from each compared method.
The averaged scores are summarized in Table~\ref{tab:quancomps}(b). Our approach consistently obtains the highest ratings across all three evaluated dimensions. These subjective preferences confirm the marked perceptual superiority of our method over existing state-of-the-art approaches.
A key observation is the discrepancy between objective metrics and user study results. This arises because objective metrics provide a compartmentalized evaluation by design, while users naturally synthesize a unified impression from all aspects of the output.

\begin{table*}[h]
\centering
\caption{(a) Quantitative comparison of style transfer performance in terms of content and style alignment. (b) User study results on three perceptual criteria: content alignment, style alignment, and overall visual quality. The most favorable outcomes are highlighted in \textbf{bold}.}
\label{tab:quancomps}

\begin{subtable}[t]{0.4\linewidth} 
\caption{ }
\centering
\setlength{\tabcolsep}{1.5mm}{
\begin{tabular}{lcc}
\toprule
Method & Content $\uparrow$ & Style $\uparrow$ \\
\midrule
Ours & 0.287 & \textbf{0.750} \\
\midrule
WSDT & \textbf{0.297} & 0.699 \\
VSP & 0.292 & 0.729 \\
StyleShot & 0.269 & 0.718 \\
StyleAligned & 0.281 & 0.700 \\
DEADiff & 0.286 & 0.687 \\
InstaStyle & 0.284 & 0.683 \\
InstantStyle & 0.280 & 0.708 \\
CSGO & 0.277 & 0.698 \\
\bottomrule
\end{tabular}}
\end{subtable}
\hfill 
\begin{subtable}[t]{0.53\linewidth} 
\caption{ }
\centering
\setlength{\tabcolsep}{1.5mm}{
\begin{tabular}{lccc}
\toprule
Method & Content $\uparrow$ & Style $\uparrow$ & Overall$\uparrow$  \\
\midrule
Ours & \textbf{4.09} & \textbf{4.31}&\textbf{4.17} \\
\midrule
WSDT & 3.74 & 4.14&   4.10\\
VSP & 3.66 &4.20 &4.03\\
StyleShot & 3.67 & 3.85&3.83 \\
StyleAligned & 3.64 & 3.71&3.72 \\
DEADiff & 3.08 &2.93 & 2.96\\
InstaStyle & 3.54 & 3.27& 3.60\\
InstantStyle & 3.55 & 3.34 &3.56\\
CSGO &3.01 & 2.79 & 2.85\\
\bottomrule
\end{tabular}}

\end{subtable}

\end{table*}

\subsection{Ablation Study}
\subsubsection{Influence of \(\kappa\).}
\begin{figure*}
\centering
\includegraphics[width=1.0\linewidth]{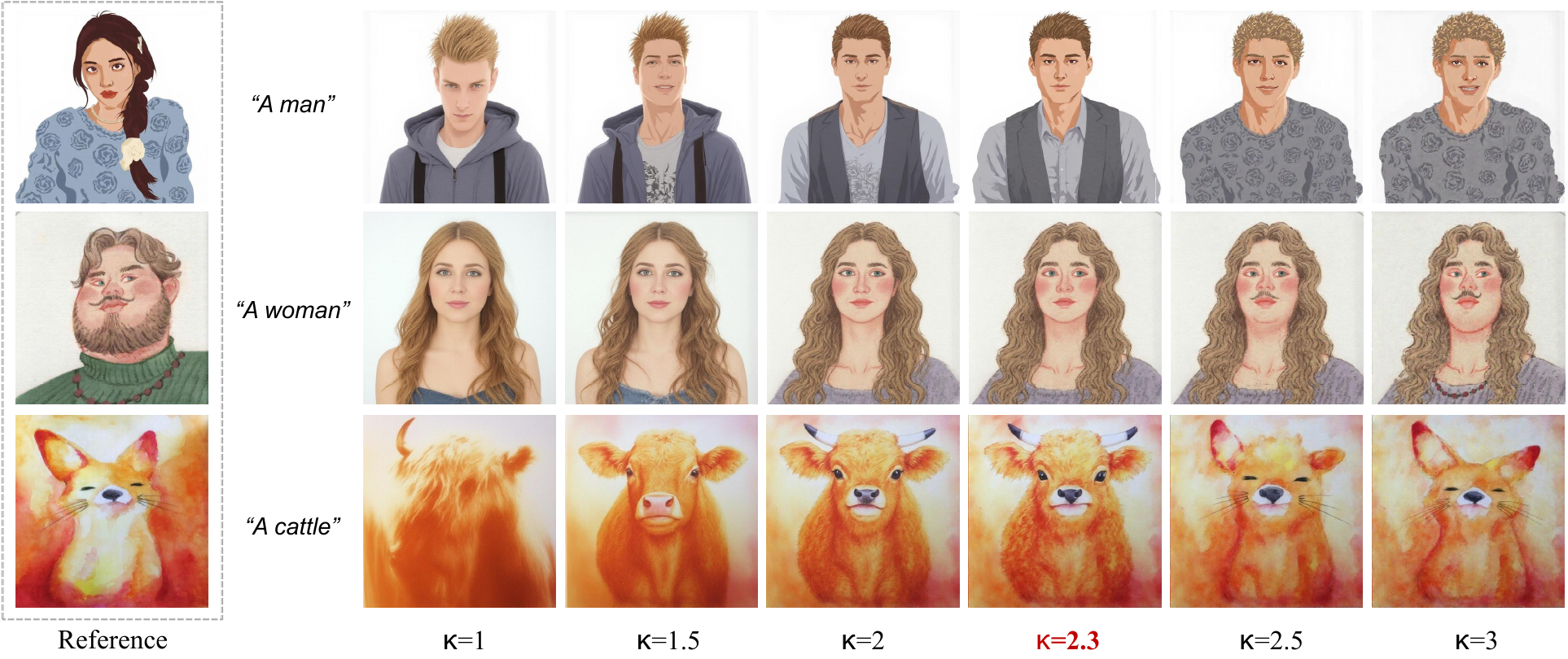}
\caption{Effects of varying \(\kappa\) on style-guided image synthesis. Smaller values of \(\kappa\) result in diminished style patterns, whereas larger values allow the semantics of the style image to influence the generated content. At \(\kappa = 2.3\), the outputs achieve an optimal balance between stylization and semantic preservation.
}
\label{fig:gamma}
\end{figure*}

To evaluate the role of the hyper-parameter $\kappa$ in our Dynamic Semantic-Style Integration (DSSI) framework, we conduct an ablation study by varying its value and assessing its impact on the generated outputs. As described in Eq.~\ref{eq:dssa}, $\kappa$ acts as a global scaling factor controlling the relative strengths of the prompt and style branches, effectively modulating the relative influence of semantic and stylistic cues during the fusion process. By adjusting $\kappa$, we can change the degree of stylization and content preservation.
We systematically vary $\kappa$ across a range of values (e.g., from 1.0 to 3.0) and generate images using diverse style references and content prompts. The results are visualized in Fig.~\ref{fig:gamma} and quantitatively summarized in Table~\ref{tab:abl}, where we report metrics such as style alignment and content alignment.
As $\kappa$ increases, the style patterns become progressively more pronounced, enhancing the visual transfer of textures, colors, and artistic elements from the reference image. For small $\kappa$ (e.g., $\kappa \leq 1.5$), the outputs exhibit subdued stylization, often resulting in images that lack distinctive artistic flair. Conversely, larger $\kappa$ (e.g., $\kappa \geq 2.5$) amplifies style dominance, potentially introducing artifacts or semantic distortions, as the reference image's features overpower the content guidance. Empirically, we find that $\kappa = 2.3$ achieves an optimal balance, yielding high-quality results where stylization is vivid yet harmonious with the semantic content. 
\begin{table*}[h]
\centering
\setlength{\tabcolsep}{2mm}{
\caption{Quantitative results demonstrating the influence of $\kappa$ on generation performance across metrics, including content alignment and style alignment. The optimal parameter is marked with (*).}
\label{tab:abl}
\begin{tabular}{l|cccccc}
\toprule
 Metric &$\kappa$=1& $\kappa$=1.5 &$\kappa$=2& $\kappa$=2.3(*)&$\kappa$=2.5&$\kappa$=3   \\
\midrule
$Content  \uparrow$ &  0.276&0.279&0.289&0.287 &  0.277 & 0.271   \\
$Style \uparrow$  &0.632 &0.646&0.700&0.750&0.756&0.760   \\
\bottomrule
\end{tabular}}
\end{table*}

\begin{figure*}
\centering
\includegraphics[width=1.0\linewidth]{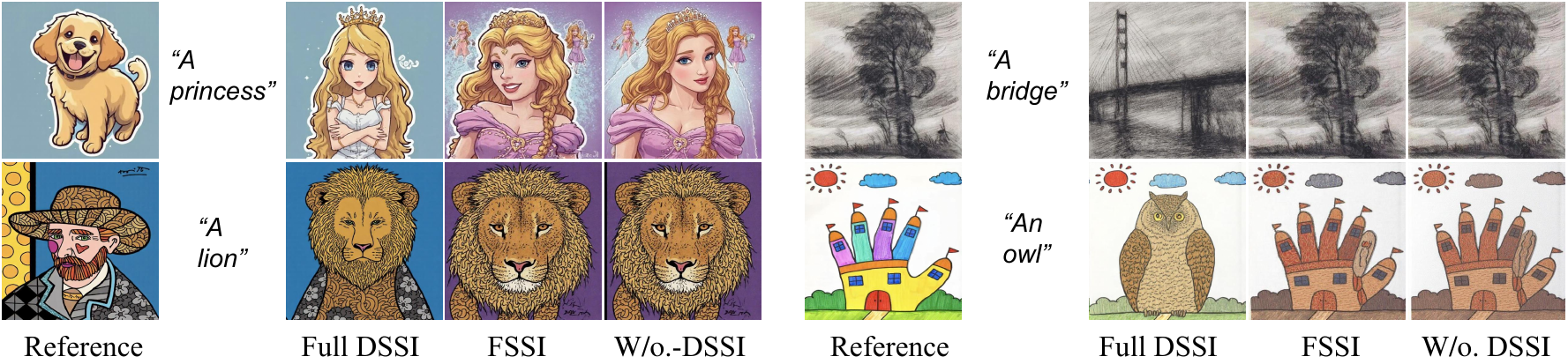}
\caption{Effects of three settings (full DSSI, simplified FSSI, and without DSSI) on style-guided image synthesis.
}
\label{fig:dssa_abl}
\end{figure*}

\subsubsection{Influence of DSSI Module.}
To assess the impact of our proposed Dynamic Semantic-Style Integration (DSSI) module on the overall performance of the framework, we conduct an ablation study by simplifying and removing the module and comparing the results across three configurations. This analysis aims to isolate the contributions of DSSI's adaptive balancing mechanism, highlighting its role in achieving robust semantic-style fusion without training.
First, we simplify the DSSI module to a fixed-weight variant (Fixed-weight Semantic-Style Integration, FSSI), where the dynamic parameter $\lambda$ is replaced with a static weight of 0.5 between the prompt and style branches:
\begin{equation}
\mathbf{h}_o^{\text{FSSI}} = \kappa\cdot\big(0.5 \cdot {\alpha}_p \mathbf{V}_p + 0.5 \cdot {\alpha}_s \mathbf{V}_s \big)+ \boldsymbol{\alpha}_o \mathbf{V}_o.
\label{eq:dssa}
\end{equation}
Second, we completely remove the DSSI module, reverting to the vanilla attention formulation:
$\mathbf{h}_o = \boldsymbol{\alpha}_p \mathbf{V}_p + \boldsymbol{\alpha}_s \mathbf{V}_s + \boldsymbol{\alpha}_o \mathbf{V}_o$.

We evaluate these three settings, including full DSSI, simplified FSSI, and without DSSI, using a diverse set of style references and content prompts, generating images under consistent hyperparameter (fixed $\kappa = 2.3$).
Qualitative results are visualized in Fig.~\ref{fig:dssa_abl}, which showcases side-by-side comparisons. 
Quantitative metrics, including style fidelity and semantic consistency, are reported in Table~\ref{tab:abl2}.

The results reveal that the full DSSI module significantly outperforms the ablated variants. Using simplified FSSI or without DSSI, outputs often exhibit noticeable imbalances: strong prompt alignments lead to weak stylization, while dominant style cues cause content distortion.
In contrast, the adaptive $\lambda$ in full DSSI dynamically adjusts to varying $\lambda_p$ and $\lambda_s$, achieving the best metrics and visually coherent results without artifacts.

\begin{table*}[h]
\centering
\setlength{\tabcolsep}{2mm}{
\caption{Quantitative ablation results for the DSSI module across metrics including content alignment and style alignment. The optimal setting is marked with (*).}
\label{tab:abl2}
\begin{tabular}{l|ccc}
\toprule
 Metric &Full DSSI(*)& Simplified FSSI & Without DSSI  \\
\midrule
$Content  \uparrow$ &  0.287&0.258&0.264   \\
$Style \uparrow$  &0.750 &0.740&0.717   \\
\bottomrule
\end{tabular}}
\end{table*}

\subsection{Application}

\subsubsection{Multiple Style-guided Image Synthesis}

\begin{figure*}
\centering
\includegraphics[width=0.9\linewidth]{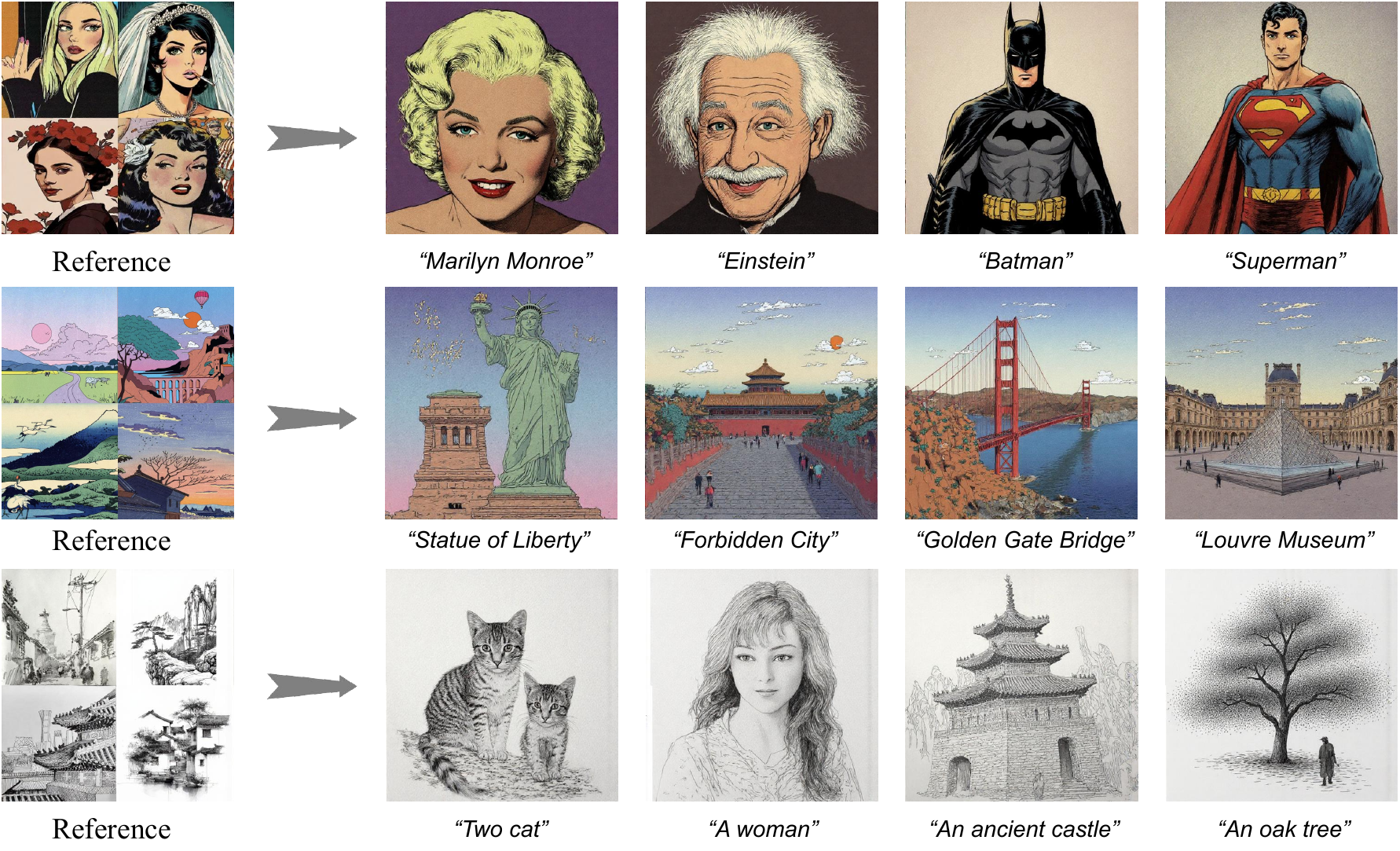}
\caption{Multiple style-guide image synthesis.
}
\label{fig:app2}
\end{figure*}

A key advantage of our approach lies in its inherent scalability to multiple style references, enabling the extraction of more robust and intrinsic stylistic representations that transcend the limitations of single-image references.
As depicted in Fig.~\ref{fig:app2}, our method seamlessly accommodates four style images by simply concatenating them.
When provided with multiple style images that share a coherent artistic theme (such as comic and sketch style), our method can distill intrinsic style representation.
The resulting stylized outputs exhibit significantly enhanced style consistency and richness, manifesting as more authentic and professionally coherent artistic renderings.

\subsubsection{Image Editing}
\begin{figure*}
\centering
\includegraphics[width=1.0\linewidth]{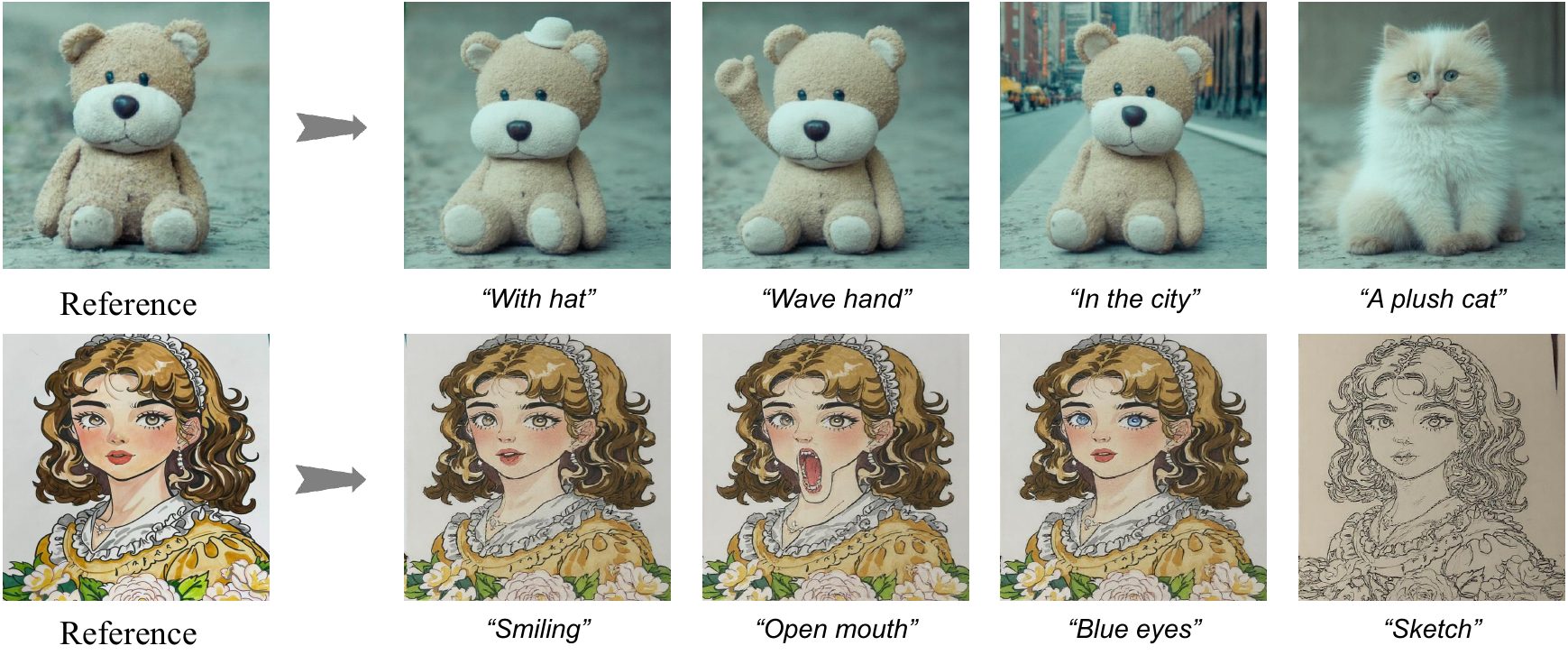}
\caption{Image editing results.
}
\label{fig:app1}
\end{figure*}

Beyond single-reference style transfer, our method naturally extends to a wide range of real-world image editing tasks, demonstrating remarkable versatility and generalization across both photorealistic and highly artistic inputs.
As illustrated in Fig.~\ref{fig:app1}, our approach seamlessly supports diverse editing operations, including adding objects, non-rigid editing, object replacement and attribute editing (e.g., color, material and style).
By changing the textual prompt, regardless of whether the input image is a natural photograph or an already heavily stylized artwork, our method adheres to the user’s textual editing instructions while preserving the content and appearance of non-edited regions.

\section{Limitation and Discussion}\label{sec12}
\begin{figure*}
\centering
\includegraphics[width=0.8\linewidth]{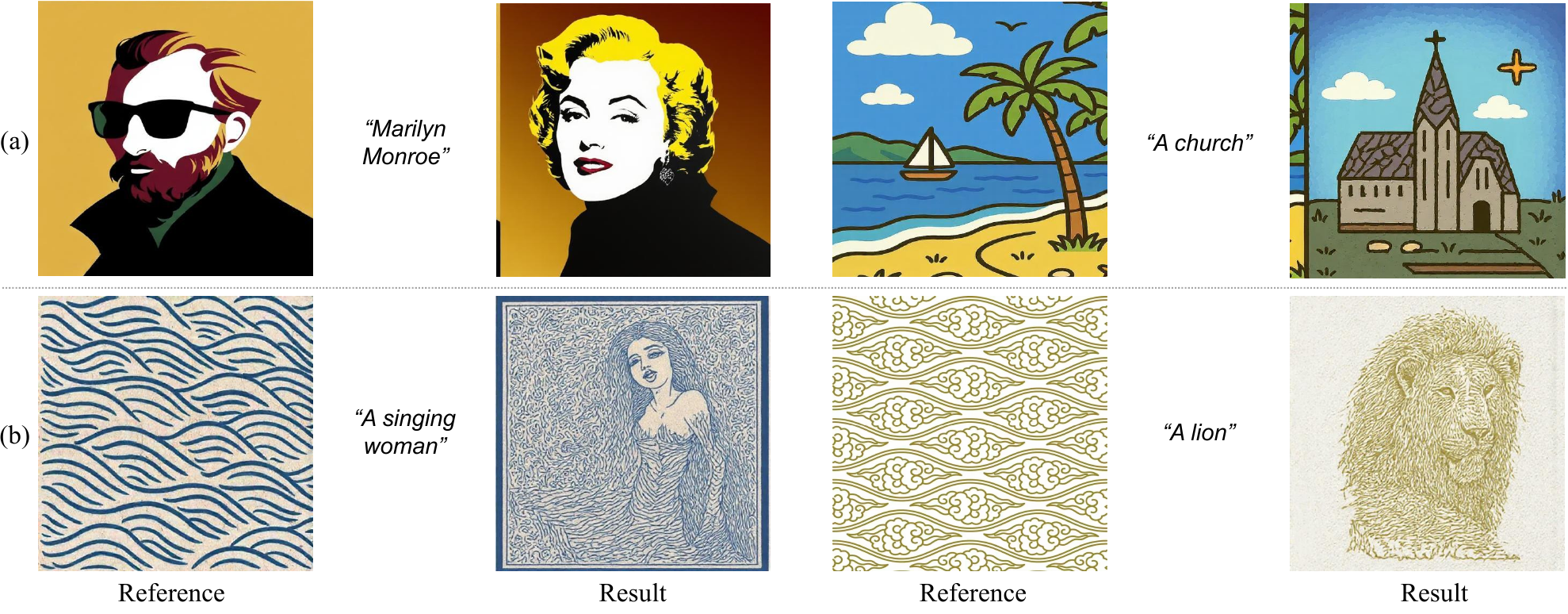}
\caption{Bad cases. (a) Inpainting Paradigm Constraint. (b) Texture Input Sensitivity.}
\label{fig:limitation}
\end{figure*}

As shown in Fig.~\ref{fig:limitation}, the limitations of Sissi can be summarized in two main aspects. (a) Due to the inpainting paradigm, the generated results may inadvertently complete or extend elements from the style reference image at the image junctions, though this does not significantly impact the overall stylized output. (b) When the input is a pure texture image, the model struggles to reasonably arrange and integrate the textures within the generated content, thereby compromising style preservation performance.
To address these limitations, future research could explore advanced masking strategies or multi-resolution fusion techniques to minimize junction artifacts in inpainting-based methods. Additionally, incorporating texture-specific priors or hierarchical attention mechanisms could improve handling of pure texture inputs, enhancing style fidelity and arrangement in diverse scenarios.

\section{Conclusion}\label{sec13}

In this paper, we introduce Sissi, a training-free style–image-guided synthesis framework that leverages the in-context capabilities of the ReFlow-based inpainting model to achieve robust and coherent semantic–style fusion. Through the proposed Dynamic Semantic–Style Integration module, our method adaptively balances textual semantic prompts with visual style references, enabling high-quality stylized generation with strong fidelity to both content and artistic attributes. Furthermore, Sissi extends naturally to a wide range of downstream applications, including multiple style-guided image synthesis, and fine-grained image editing, demonstrating its versatility and effectiveness across diverse generative tasks.

\bibliography{sn-bibliography}

\end{document}